\documentclass[10pt,twocolumn]{article} 
\usepackage{simpleConference}

\usepackage{times}
\usepackage{epsfig}
\usepackage{graphicx}
\usepackage{amsmath}
\usepackage{amssymb}
\usepackage{comment}
\usepackage{csquotes}
\usepackage{booktabs}
\usepackage{hhline}
\usepackage{floatrow}
\usepackage{subcaption}
\usepackage{placeins}
\usepackage{url}

\usepackage[pagebackref=true,breaklinks=true,letterpaper=true,colorlinks,bookmarks=false]{hyperref}
\usepackage{cleveref}

\begin{document}

\title{ELFIS: Expert Learning for \\ Fine-grained Image Recognition Using Subsets}

\author{
Pablo Villacorta$^*$, 
Jesús M. Rodríguez-de-Vera$^*$, 
Marc Bolaños,\\
Ignacio Sarasúa, 
Bhalaji Nagarajan, and 
Petia Radeva \\
}
\maketitle
\def\thefootnote{*}\footnotetext{These authors contributed equally to this work}
\thispagestyle{empty}

\begin{abstract}
Fine-Grained Visual Recognition (FGVR) tackles the problem of distinguishing highly similar categories. One of the main approaches to FGVR, namely subset learning, tries to leverage information from existing class taxonomies to improve the performance of deep neural networks. However, these methods rely on the existence of handcrafted hierarchies that are not necessarily optimal for the models. In this paper, we propose ELFIS, an expert learning framework for FGVR that clusters categories of the dataset into meta-categories using both dataset-inherent lexical and model-specific information. A set of neural networks-based experts are trained focusing on the meta-categories and are integrated into a multi-task framework. 
Extensive experimentation shows improvements in the SoTA FGVR benchmarks of up to +1.3\% of accuracy using both CNNs and transformer-based networks. Overall, the obtained results evidence that ELFIS can be applied on top of any classification model, enabling the obtention of SoTA results. 
The source code will be made public soon.
\end{abstract}

\section{Introduction}
The advent of Deep Learning (DL)
\cite{726791} has led to a super-human performance in many tasks such as image classification \cite{he2016deep, liu2022convnet, tan2019efficientnet, tan2021efficientnetv2}. 
However, classification is still an open field in case of a large number of classes where usually groups of classes with significant confusion occur \cite{lin2017end, tsoumakas2008effective}. Fine-Grained Visual Recognition (FGVR) is an image classification problem that aims at recognizing images that belong to different subordinate categories (analogously to the target classes in general classification tasks) of a meta-category. In contrast to general image classification tasks, FGVR presents some particular challenges. Fine-grained visual classification images 
have more detailed categories with subtle differences (low inter-class variance), making it harder to distinguish between
categories \cite{wei2021fine}. On the other hand, pose, illumination, and background often contribute to large differences among objects in the same category (large intra-class variance) \cite{zhao2017survey}. 

In order to address these issues, different approaches have been proposed in the literature. Typically, data-driven techniques are followed, leaving it to the neural networks to discover subtle discriminatory details between the different subordinate classes  \cite{cap, 7299194, huang2020interpretable, dubey2018pairwise}. 
Popular and well-established FGVR approaches fall into either the use of localized part-based descriptors \cite{wei2021fine, wang2018learning, huang2020interpretable, zeiler2014visualizing} or end-to-end feature learning \cite{dubey2018pairwise, dubey2018maximum, cap}.
A less explored FGVR paradigm is
\textit{subset learning}, which focuses on exploiting prior information during training by leveraging explicit hierarchies of categories \cite{chang2021your}. Recently proposed methods that fall under this strategy group 
require constructing multi-level hierarchies that are model-agnostic \cite{7350861, chang2021your, ijcai2021p100}. For instance, the work \cite{ijcai2021p100} introduces a method to combine the outputs of different subset-specific networks, where each subset is extracted from an automatically generated hierarchy of classes derived from an external knowledge base.
Subset learning methods have shown strong classification performance in complex real-world domains 
like food recognition \cite{rodenas2022learning}. Using a per-class subnetwork expert, the authors of \cite{rodenas2022learning} propose to force the network to focus on the most challenging classes based on the confusion matrix of a baseline. 



\noindent \textbf{Our proposal:} In this paper, we rethink the problem of subset-based fine-grained recognition and propose a new \textbf{E}xpert \textbf{L}earning framework for \textbf{F}ine-grained \textbf{I}mage Recognition using \textbf{S}ubset information (ELFIS). In order to construct the meta-categories (subsets of classes), we introduce a new dissimilarity metric that is based on the lexical similarity of the class labels. However, to take into account different classes with similar visual appearance, we combine the lexicographic distance with a distance expressing the confusion of a pre-trained model.
Once the subsets are constructed, a set of experts (convolutional or transformer blocks) receive a task to classify the classes from a specific subset and they are all jointly trained in a multi-task setting and aggregated to classify the whole set of classes.
Hence, each expert in charge of classifying elements within a given subset of classes contributes and shares its knowledge with others, helping the network to distinguish the classes better.

Note that thanks to combining the two sources of information, we can take into account dataset-related information (lexical) as well as the main weaknesses of the network, derived from its confusion matrix. Thus, we provide the network with richer information during training, which is performed in an online multi-task fashion, avoiding complex pipelines. We find that this improved clustering proposal not only avoids the necessity of handcrafted taxonomies for subset learning but also outperforms the results obtained when using logical or manually created groups. 

\noindent In this paper, \textbf{our contributions}  are the following: 
\begin{enumerate}
    \item A simple, automatic, intuitive, and interpretable way of combining classes in subsets based on a new distance that integrates lexical class labels information with visual class similarities derived from model confusion.
    \item A new multi-task framework that integrates experts specialized in the different subsets and a generic classifier to share and complete their knowledge to achieve high-performing FGVR. Our model is generic, i.e., it can be applied on any classifier backbone (CNN or transformer), improving the fine-grained classification accuracy relying on the aforementioned subsets.
    \item An extensive analysis of the different components of ELFIS  on 5 FGVR datasets and with 4 different backbones (2 CNNs and 2 transformers). Thanks to its versatility, ELFIS outperforms the SoTA in 2 public datasets (CUB-200-2011 and Food-101) and achieves competitive results in the other 3 public FGVR datasets.
\end{enumerate}

\section{Related Work}

\noindent \textbf{Fine-grained visual recognition.} Most  methods for FGVR can be grouped into three main paradigms: recognition by 1) localization-classification subnetworks \cite{girshick2014rich}, 2) end-to-end feature encoding \cite{dubey2018pairwise}, and 3) subset learning \cite{7350861}.
Localization-classification subnetworks are a historically popular approach, where a localization submodule first locates the key parts of an object and generates a local part-level feature vector for each of the parts prior to the final classification \cite{7410088}. Different approaches vary in how the localization subnetwork is modeled, such as employing detection or segmentation techniques \cite{girshick2014rich, 7299194, He_Peng_2017} or utilizing activations from convolutional layers \cite{simon2015neural, 7780497, 9008286}. 


Recognition by end-to-end feature encoding is a paradigm of FGVR that focuses on feature learning \cite{wei2021fine}. Different methods have been developed to learn discriminative representations using approaches like high-order feature interactions \cite{dubey2018pairwise} or designing specific loss functions \cite{dubey2018maximum}. 
Subpixel gradients are used to detect subtle variations and differentiate fine-grained subcategories using Context-aware Attentional Pooling \cite{cap}. SnapMix \cite{huang2021snapmix} presents a semantically proportional mixing that uses Class Activation Mapping \cite{zhou2016learning} to minimize label noise during training.

\noindent \textbf{Subset learning} is a less explored group of approaches in which the problem of FGVR is tackled by extracting subsets of similar classes and training an expert model for each subset. The work \cite{7350861} introduces a hierarchical subset learning approach where the images of a dataset are clustered based on deep convolutional features. For each subset, they train a local SVM classifier that discriminates between the classes of its subset, and they also train classifiers to act as \textit{subset selectors} during inference. The authors of \cite{ge2015subset} introduce a similar approach, but learning features specific to each subset using separate subset-specific CNNs.
\cite{chang2021your} propose a framework that performs multi-granularity label predictions, building on top of a manually generated hierarchical taxonomy of birds, based on information extracted from Wikipedia. DeepMe \cite{ijcai2021p100} was proposed as a method to combine the outputs of different networks based on different subsets of classes, obtained by applying a novel soft spectral clustering procedure to build a two-layer ontology, based on the semantic distances between the classes in the WordNet hierarchy \cite{10.1145/219717.219748}.
In \cite{rodenas2022learning}, subsets of classes are extracted based on model errors, and the different experts are modeled as parallel prediction heads within a single model. 
The method requires an expensive 3-stage training pipeline, as well as a sophisticated re-sampling method that severely affects training time. 

\begin{figure*}
\begin{center}
\includegraphics[width=0.96\textwidth,trim=0 0 0 0,clip]{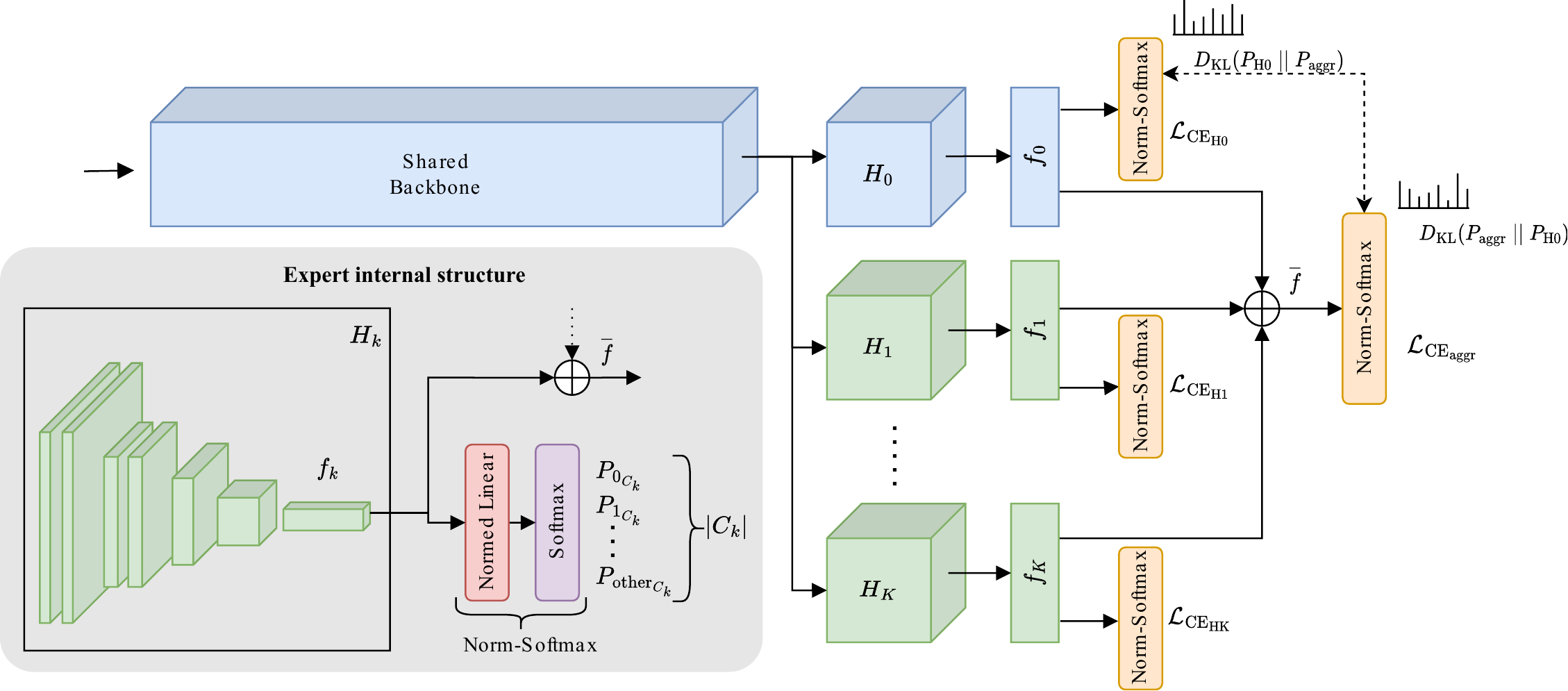}
\end{center}
   \caption{ELFIS framework. The blue blocks of the model correspond to the baseline architecture, while the green blocks represent the subset experts or classification heads. The gray box contains the structure of each one of the experts, which is formed by a series of convolutional or transformer layers and a normed linear final classifier.}
\label{fig:diagram}
\end{figure*}

\noindent \textbf{Knowledge distillation and mutual learning.} Knowledge distillation is a subfield of DL that has received rapidly increasing attention from the community \cite{gou2021knowledge}. More recently, mutual learning has emerged as a generalization of knowledge distillation where multiple models jointly learn from each other \cite{zhang2018deep, guo2020online}. In \cite{chen2020online}, authors perform two-level distillation during training with multiple auxiliary peers and one group leader that leads the learning process. \cite{Park_2023_WACV} shows the benefits of using mutual learning between different experts  trained with subsets of the data. They use KL-divergence loss between the different probability distributions of the expert networks, finding more robust minima and acquiring better representations.
In \cite{KE2023109305}, the authors introduce GDSMP-Net, a framework adopting a multi-granularity hierarchical feature fusion learning strategy that uses cross-layer self-distillation to improve the robustness of the learned multi-granularity features. However, the usage of mutual learning for FGVR is mainly unexplored. In this paper, we propose utilizing two-way knowledge distillation to boost the performance of both the main classifier and the aggregation layer, using distillation only at the category level.

\section{Expert Learning for FGVR Using Subsets}

In this section, we introduce ELFIS, a universal add-on that enhances the fine-grained recognition capabilities of any given base image classifier. \Cref{fig:diagram} illustrates our proposed framework, outlining its main components.

Our proposal is built around the idea of identifying subsets of classes that are hard to distinguish. The method starts by training a classification network, the \textit{baseline}, for the target dataset. This network, which can be a CNN or vision transformer, is represented in blue in \Cref{fig:diagram}. The confusion matrix of the baseline is used as an assessment of the weaknesses of the model, and this information is combined with lexical information from the dataset labels to eventually find similar classes and group them together into multiple subsets. This process is detailed in \Cref{sec:proposal-subset-construction}.

As we describe in \Cref{sec:subset-heads}, once the subsets have been identified, we add to the original architecture as many subnetwork experts as there are subsets (shown in green in \Cref{fig:diagram}) parallel to the original classification head. Moreover, a final aggregation layer (right-most element in \Cref{fig:diagram}) is introduced to combine the information from the different experts into a final prediction (see \Cref{sec:final-prediction}). The model is then trained in an online multi-task learning fashion (\Cref{sec:proposal-mutual,sec:proposal-training}), resulting in an improved performance in fine-grained classification with respect to the baseline.

\subsection{Baseline architecture} 

Our baseline architecture is divided into a backbone and a final classification head, $H_0$, and it is trained on the whole set of classes. 

The backbone receives as input the original image and outputs its representation in the feature space. 
At the end of $H_0$, a linear layer outputs $n$ logits which are then passed through a softmax operator to obtain the class-wise probabilities predicted by the network. The input of this last linear layer is $f_0\in\mathbb{R}^d$, which is the last feature representation vector of the image within the baseline network.

\subsection{Subset construction} \label{sec:proposal-subset-construction}

In this paper, we introduce a new distance to group similar classes into optimal class subsets 
by utilizing two types of information: lexical, which is model-agnostic, and visual, which is delivered by the baseline model's difficulty to distinguish similar classes.

\textbf{Lexical similarity.} Concepts are compared on the basis of the similarity of their class labels. For this purpose, the Universal Sentence Encoder \cite{cer2018universal} is used to obtain a feature representation $E_i$ of each class name, $c_i$ with $1\leq i\leq n$. It is then used to define the lexical dissimilarity of any two classes as follows:
\begin{equation}
\label{eq:lex-sim}
\operatorname{dis_{lex}}(c_i, c_j) = 1 - \frac{E_i \cdot E_j}{\|E_i\|\|E_j\|},
\end{equation}
where the second term corresponds to the cosine similarity of $E_i$ and $E_j$. Once we perform this computation for each pair of classes in the vocabulary, we obtain a dissimilarity matrix $D_{LEX}\in \mathbb{R}^{n\times n}$, where $n$ is the number of classes.



\textbf{Visual similarity.} Inspired by \cite{rodenas2022learning}, we utilize the baseline model confusion matrix to encode knowledge regarding which classes are hard to distinguish by a specific neural network. More concretely, let us consider that $M\in \mathbb{R}^{n\times n}$ is the row-wise normalized confusion matrix of a baseline. The larger $M_{ij}$ is, the more similar $c_i$ and $c_j$ are for the model. In this way, we define:
\begin{equation*}
    D_{CM} = 1 - \frac{1}{2}(M +  M^{\top})
\end{equation*}
as the matrix that stores the distance between any pair of concepts $c_i,c_j$ according to the network's representations.

\textbf{Aggregation.} A hierarchical agglomerative clustering (HAC) technique is used in order to construct the subsets of categories based on the sum of lexical and visual distances. Applying HAC to a set of classes using more than one pairwise dissimilarity value is a non-trivial problem \cite{chavent2018clustgeo}. Moreover, although the theoretical range of $D_{LEX}$ and $D_{CM}$ is very similar, their composition and nature are very different. In FGVR, where there is usually a large number of classes, there is no confusion between many of the pairs of classes. Moreover, if the baseline is already performing well, the values of $M$ outside of the diagonal will be very low. For that reason, most of the values of $D_{CM}$ are either 1 or very close to 1, making the \enquote{relevant} values very sparse and close to the maximum. On the other hand, the values of $D_{LEX}$ are much less sparse, as they are derived from the cosine similarity between real-valued vectors.

Given these differences, we first perform Z-score standardization of $D_{CM}$ and $D_{LEX}$, so that they have the same mean and standard deviation. We then average them to synthesize the information from both dissimilarity matrices into a single matrix $\overline{D}$:
\begin{equation}\label{eq:avg-std}
    \begin{aligned}
        \overline{D} &= \frac{1}{2}(\overline{D}_{CM} + \overline{D}_{LEX}),
    \end{aligned}
\end{equation}
where $\overline{D}_{CM} = \operatorname{standardize}(D_{CM})$ and $\overline{D}_{LEX} = \operatorname{standardize}$.
Finally, a HAC process using single-linkage is applied using $\overline{D}$ as the input dissimilarity matrix. Despite the standardization process, the values of $\overline{D}_{CM}$ are still very sparse. For that reason, the lexical dimension provides the main structure of the clusters, giving relevant information about all the classes, while $D_{CM}$ helps to pull together highly confused classes in the clustering space. In this way, the network receives additional guidance for all the classes but is focused primarily on separating classes that are difficult to distinguish.

\subsection{Subset heads}\label{sec:subset-heads}


For each one of the subsets $\mathcal{C}_k$, a new classification head $H_k$ is added at the end of the backbone, representing an expert specialized in its subset of classes. Here, the backbone acts as a common image encoder for all the subset heads.
$H_k$ is composed of a series of convolutional or transformer blocks, followed by a normed linear layer \cite{wang2020frustratingly}. The former is responsible for producing a feature representation $f_k\in\mathbb{R}^{d}$ that better describes the differences between the classes of that subset. The latter classifies every image into $|\mathcal{C}_k|+1$, that is, the specific class within its corresponding subset or \enquote{other} whenever the concept of the picture belongs to a different subset. 

A significant imbalance is to be expected in each one of these heads: Most of the examples seen during training will belong to the \enquote{other} class. Consequently, we propose to replace the traditional dot product of linear layers with the following expression to compute the logit value for the $i$-th class of the $k$-th head:
\begin{equation*}
    s^k_i = \frac{f_k^{\top}w_i}{\|f_k\|\|w_i\|},
\end{equation*}
where $w_i$ is the $i$-th column of the weight matrix of the normed linear layer. This is a simple mechanism that has proven to be useful in object detection, where the class \enquote{background} also accounts for most of the examples \cite{wang2020frustratingly}. By using this approach, we avoid the usage of complex weighting or balancing techniques. It is important to note that, in general, the $i$-th class of $H_k$ does not correspond to the $i$-th class in the original vocabulary of the dataset. Apart from the last classification layer, the architecture of the original head $H_0$ and the subset heads or experts is the same.

\subsection{Final prediction} \label{sec:final-prediction}

In order to make a final prediction for an input image, the classification scores of the different heads are disregarded, and only the intermediate feature representations $f_0,f_1,\dots,f_K$ are used. More specifically, the mean vector of those $K+1$ representations is computed as follows:
\begin{equation*}
    \overline{f} = \frac{1}{K+1}\sum_{0\leq k \leq K+1} f_k, 
\end{equation*}
where $\overline{f}$ encodes relevant information to better distinguish the elements within the same cluster, and the average operation acts as a regularizer. The original $H_0$, which is trained to predict the specific category of the image like in the baseline model, produces a feature vector $f_0$ which is refined with specific per-cluster information from the different subset heads. $\overline{f}$ is then fed into a linear layer with $n$ outputs, which is used to produce the final output of the model.  

\subsection{Mutual learning to boost aggregation quality} \label{sec:proposal-mutual}

In addition to the traditional classification loss, we add a mutual learning loss component that maximizes the mutual information between the probability distribution output by $H_0$ and by the aggregation layer. More concretely, we minimize the Kullback-Leibler divergence \cite{kullback1951information} between both probability distributions for each input image:
\begin{equation}\label{eq:mutual-learning}
\begin{aligned}
    \mathcal{L}_{ML_{H0}} &= D_{KL}(P_{H0}\|P_{aggr}) \\
    \mathcal{L}_{ML_{aggr}} &= D_{KL}(P_{aggr}\| P_{H0}).
\end{aligned}
\end{equation}

The reason for adding $D_{KL}(P_{H0}\|P_{aggr})$ is to distill the improved performance of the  different heads aggregation to the original head $H_0$: The better the input of the aggregation layer, the better the results we will obtain from it.
On the other hand, we also include $D_{KL}(P_{aggr}\| P_{H0})$ to provide additional guidance to the learning process of the aggregation layer and ensure that the performance of the aggregation is at least as good as that of $H_0$, as well as to mitigate the potentially detrimental effect of noisy signals. Results reported in \Cref{sec:ablations} show empirically the benefits of this two-way distillation.

\subsection{Training} \label{sec:proposal-training}

The backbone and the different heads or experts are trained jointly, and the optimization criterion is a loss composed of $2+K+2$ terms organized into 3 main components:
\begin{equation*}
\begin{aligned}
    \mathcal{L} =&\lambda_1 (\mathcal{L_{\mathrm{CE}_{\mathrm{H0}}}} + \mathcal{L}_{ML_{H0}}) 
    +\lambda_2 \frac{1}{K} \sum_{k=1}^K \mathcal{L_{\mathrm{CE}_{\mathrm{Hk}}}} \\
    &+\lambda_3 (\mathcal{L_{\mathrm{CE}_{\mathrm{aggr}}}}+\mathcal{L}_{ML_{aggr}}),
\end{aligned}
\end{equation*}
where $\mathcal{L}_{\mathrm{CE}}$ represents the cross-entropy loss and $\lambda_1, \lambda_2, \lambda_3 > 0$ weigh the contribution of the loss of the original classification head, the subset heads, and the aggregation layer, respectively. We empirically set the values $\lambda_1=1,\lambda_2=10,\lambda_3=1$ for all the experiments. The gradients flow throughout the whole network, which forces the experts to learn not only to produce features that are useful for the subproblem of distinguishing the classes of a single subset but also features that are useful for the final aggregation. Moreover, this multi-task setting also enables additional information to flow to the different heads and the backbone, which enriches the training process.

\section{Experiments} \label{sec:experiments}
In this section, we discuss the implementation details, the datasets, quantitative and qualitative results including the ablation study, and comparison to the SoTA.

\subsection{Implementation details}

We tested our proposal, ELFIS, with four different backbones. Two of them are small-sized CNNs (EfficientNet-B0 \cite{tan2019efficientnet} and ResNet-50 \cite{he2016deep}) and the other two are larger vision transformers (ViT-B/16 \cite{dosovitskiy2020image} and SwinV2-B \cite{liu2022swin}). We initialize both CNNs with the pre-trained weights for Imagenet-1K \cite{deng2009imagenet}, while for the transformers we used the weights pre-trained with Imagenet-21K \cite{deng2009imagenet}.

In all the experiments, we use the one-cycle learning rate scheduler proposed in \cite{smith2019super}, with a cosine annealing cycle of 40 epochs and an annihilation phase of 160 epochs. An early-stop mechanism with a patience of 15 epochs was set to prevent overfitting. The maximum learning rate was tuned for each architecture to optimize the results of the baseline on CUB-200-2011 \cite{wah2011caltech}. The ELFIS hyperparameters were kept untouched to avoid  potential bias during the evaluation and analysis. We use SnapMix \cite{huang2021snapmix}, a SoTA FGVR augmentation technique, in both the baseline and the ELFIS experiments with a probability of 0.5. Unless stated otherwise, we used images of size $224 \times 224$ for all the experiments. We run all experiments on a single NVIDIA GeForce RTX 3090 GPU, except for 
transformer-based architectures
on Food-101, where we use 8xNVIDIA V100 GPUs.


\noindent \textbf{Datasets} A total of 5 public datasets of varying domains and sizes were used for the experimentation: CUB-200-2011 \cite{wah2011caltech}, Stanford Cars \cite{krause20133d}, FGVC-Aircraft \cite{maji2013fine}, Food-101 \cite{bossard2014food} and NABirds \cite{van2015building}. For each dataset, we take as many clusters as $10\%$ of the number of categories, based on the experimentation explained in \Cref{sec:ablations}. The only exception is NABirds, for which we decided to use a lower number of clusters (25 instead of 55) to keep the number of experts low. No additional training images were used.

\subsection{Quantitative results}


We report in \Cref{tab:results-backbones} the accuracy achieved by ELFIS on four different backbone model architectures on two datasets of 
different domains and sizes: CUB-200-2011 and Food-101. We compare the results obtained by ELFIS with those obtained with the corresponding baseline model. We show that the use of ELFIS leads considerable increase in performance for all backbones, even in the case of SwinV2-B, in which the baselines already present very strong performance. 
Furthermore, the improvements remain consistent across different datasets, hinting that our method generalizes well regardless of the characteristics of the data.

We further extend these results in \Cref{tab:effnetb0}, which contains the accuracy achieved in all five considered datasets when using EfficientNet-B0 as the backbone and an image size of $384\times 384$. As can be seen, applying ELFIS leads to clearly noticeable improvements in almost all datasets. The performance gain is especially high in the case of NABirds, with an improvement of $+1.18\%$ accuracy points.

\begin{table}[t]
\centering
\begin{tabular}{@{}llrr@{}}
\toprule
Method   & Backbone        & \multicolumn{1}{c}{CUB} & \multicolumn{1}{c}{Food-101} \\ \midrule
Baseline & EfficientNet-B0 & 85.85                   & 91.22                        \\
ELFIS    & EfficientNet-B0 & \textbf{86.32}          & \textbf{91.23}               \\ \midrule
Baseline & ResNet-50       & 86.47                   & 90.23                        \\
ELFIS    & ResNet-50       & \textbf{87.94}          & \textbf{91.38}               \\ \midrule
Baseline & ViT-B/16        & 89.85                   & 92.22                        \\
ELFIS    & ViT-B/16        & \textbf{90.45}          & \textbf{93.09}                    \\ \midrule
Baseline & SwinV2-B        & 92.08                   & 94.81                        \\
ELFIS    & SwinV2-B        & \textbf{92.22}          & \textbf{95.05}                    \\ \bottomrule
\end{tabular}
\caption{Results of ELFIS in CUB-200-2011 and Food-101 with different backbones.}
\label{tab:results-backbones}
\end{table}

\begin{table}[t]
\centering
\begin{tabular}{@{}llccccc@{}}
\toprule
Method        & CUB   & Cars   & Aircraft & Food-101 & Birds \\ \midrule
Baseline  & 85.85 & 94.18 & 90.19   & 91.22  & 81.54   \\
ELFIS     & \textbf{86.32} & \textbf{94.48}  & \textbf{90.89}    & \textbf{91.23}   &  \textbf{82.72}       \\ \bottomrule
\end{tabular}
\caption{
Results across datasets with EfficientNet-B0.
}
\label{tab:effnetb0}
\end{table}

We also find that, as a result of taking into account the confusion matrix of the baseline to build the subsets, ELFIS enhances the network by focusing more on the most difficult classes, leading to a class-wise accuracy improvement for most of the classes. In terms of class-wise classification accuracy, ELFIS tends to clearly outperform the baseline in the vast majority of the dataset categories, and in the few cases where the results worsen, the performance of ELFIS is still very close to that of the baseline. We observe this pattern in almost all of the configurations we studied, confirming the ability of ELFIS to not only match but also improve the performance of the original baseline.

A more detailed analysis of the accuracy improvements introduced by ELFIS can be found in Appendix A.

\subsubsection{Comparison to the state-of-the-art}

We report the SoTA results for both CUB and Food-101 in \Cref{tab:sota}, and compare them with those obtained using ELFIS. We achieve SoTA performance for the Food-101 dataset (95.1\%), and competitive results in CUB-200-2011 (92.2\%). Note that our results would correspond to ranking second in the Papers With Code CUB FGVR benchmark.
As previously explained, ELFIS also provides great improvements in smaller networks. 
EfficientNet-B0 is a much smaller backbone than those used in the literature. Even in that case, results reported in \Cref{tab:effnetb0} are close to SoTA solutions in other datasets. For example, our results with that smallest network are only 1\% below the current best solution for Stanford Cars - CAP (95.7\%)  \cite{cap}. Most of the SoTA methods also use bigger image sizes, further distorting the real performance gap with ELFIS. Our method improves upon the highest-performing cases on the datasets where we used bigger backbones, and we can expect the trend to continue and improve current SoTA methods if applied on top of them.
On the whole, the comparison from \Cref{tab:sota} suggests that ELFIS is not only capable of  improving with respect to its corresponding baseline, but also that these improvements are on par with the current SoTA approaches.    


\begin{table}[t]
\centering
\begin{tabular}{@{}lcc@{}}
\toprule
Method                 & CUB & Food-101   \\ \midrule
ViT-Net \cite{pmlr-v162-kim22g} (ICML'22)$^{\dag}$   & 91.7 & -  \\
CAP \cite{cap} (AAAI'21)        & 91.8  &  -   \\
DCAL \cite{zhu2022dual} (CVPR'22)$^{\dag}$      & 91.4 &   -   \\
Grafit \cite{touvron2021grafit} (ICCV'21)               & -    & 93.7 \\
EffNet-B7 \cite{tan2019efficientnet} (ICML'19)$^{\dag}$  & - & 93.0 \\
PMG \cite{chang2021your} (CVPR'21)$^{\dag\S}$      & 89.9 &   87.5   \\
FGFR \cite{rodenas2022learning} (Madima'22)$^{\S}$   & 90.4 &  93.8   \\
\midrule
ELFIS + SwinV2-B$^{\S}$  & \textbf{92.2} & \textbf{95.1}  \\ \bottomrule
\end{tabular}
\caption{Comparison of the proposed ELFIS with SoTA methods. $\dag$ denotes methods using a bigger image size than ours. $\S$ is used to distinguish subset-based methods.}
\label{tab:sota}
\end{table}

\subsubsection{Other ablations} \label{sec:ablations}

In this section, we conduct a series of ablation experiments to study the effect of various hyperparameters and design choices of our proposed method. Through these experiments, we aim to demonstrate the effectiveness and robustness of our method, as well as to provide insights into its underlying mechanisms.
We perform an analysis of the behavior of the following aspects of ELFIS: the head output aggregation method, the cluster construction technique, the number of clusters to generate, the use of distillation, as well as the size of each of the model experts. For each of these experiments, we build on top of the best configuration reported in the previous ablation.

\begin{table*}
\centering
\begin{subtable}[b]{0.275\linewidth}
\centering
\begin{tabular}{@{}lc@{}}
\toprule
Aggregation method & ACC  \\ \midrule
Mean               & \textbf{94.1} \\
1D convolution     & 93.7 \\
Concatenation      & 93.6 \\
Median             & 93.5 \\
Max                & 94.0 \\ \bottomrule
\end{tabular}
\vspace{0.5em}
\caption{Aggregation of features on Cars.}
\label{tab:ablation-aggregation}
\end{subtable}%
\begin{subtable}[b]{0.45\linewidth}
\centering
\begin{tabular}{@{}ccc|ll@{}}
\toprule
\multicolumn{1}{l}{CM} & \multicolumn{1}{l}{LEX} & \multicolumn{1}{l|}{Combination} & Cars & CUB \\ \midrule
        -               &         -                & -                                &   93.6   &   82.8  \\
$\checkmark$            &       -                  & -                                &    94.0  &   \textbf{83.8}  \\
       -                & $\checkmark$             & -                                &   94.0   &   83.4  \\
$\checkmark$            & $\checkmark$             & Average                          &   93.7   &   83.3  \\
$\checkmark$            & $\checkmark$             & Std Average                      &   \textbf{94.1}   &   \textbf{83.8}  \\
$\checkmark$            & $\checkmark$             & ClustGeo                         &   93.7   &   83.6   \\ \bottomrule
\end{tabular}
\caption{Cluster construction approaches.}
\label{tab:ablation-cluster-construction}
\end{subtable}%
\begin{subtable}[b]{0.225\linewidth}
\centering
\begin{tabular}{@{}lc@{}}
\toprule
             & ACC           \\ \midrule
Baseline     & 90.2          \\
Manufacturer & 90.2          \\
Proposed     & \textbf{90.9} \\ \bottomrule
\end{tabular}
\vspace{1.75em}
\caption{Results on FGVC-Aircraft.}
\label{tab:ablation-aircraft}
\end{subtable}
\vspace{0.75em}

\begin{subtable}{0.235\linewidth}
\centering
\begin{tabular}{@{}lccc@{}}
\toprule
     & 5\%  & 10\% & 15\% \\ \midrule
CUB  & 83.4 & \textbf{83.7} & \textbf{83.7} \\
Cars & 93.9 & \textbf{94.1} & 93.9 \\ \bottomrule
\end{tabular}
\vspace{0.75em}
\caption{Number of clusters.}
\label{tab:ablation-num-clusters}
\end{subtable}%
\begin{subtable}{0.5\linewidth}
\centering
\begin{tabular}{@{}cc|ll@{}}
\toprule
$KL(P_{H0}\|P_{aggr})$ & $KL(P_{aggr}\| P_{H0})$ & Cars & CUB \\ \midrule
                       &                         &   \textbf{94.1}   &  83.7   \\
$\checkmark$           &                         &   \textbf{94.1}   &  83.2   \\
$\checkmark$           & $\checkmark$            &   \textbf{94.1}   &  \textbf{84.3}   \\ \bottomrule
\end{tabular}
\caption{Distillation settings.}
\label{tab:ablation-mutual}
\end{subtable}%
\begin{subtable}{0.25\linewidth}
\centering
\begin{tabular}{@{}lccc@{}}
\toprule
     & \multicolumn{3}{c}{Blocks} \\ 
     & 0            & 1               & 2           \\ \midrule
Cars &         93.9     & \textbf{94.1}            &        93.5     \\ \bottomrule
\end{tabular}
\vspace{0.75em}
\caption{Number of blocks.}
\label{tab:ablation-num-blocks}
\end{subtable}
\caption{Ablations of the different components of ELFIS. All results are obtained using EfficientNet-B0.}
\end{table*}

\noindent \textbf{Head aggregation.}
In order to generate an aggregated prediction, we explore multiple ways of combining the intermediate feature representations produced by the experts and report the results for Stanford Cars in \Cref{tab:ablation-aggregation}. We find that computing the average of the feature vectors yields the best performance from the method, while also being computationally cheap compared to other approaches such as concatenating the representations into a single vector.


\noindent \textbf{Cluster construction.} We experiment with a variety of approaches with which to obtain the class clusters that ELFIS builds on top of. We explore using only either the confusion matrix- and lexical-based distances separately, as well as different approaches to combine them into a single distance matrix. In addition to Eq. \eqref{eq:avg-std}, we also tested the non-standardized average of the dissimilarity matrices and the compound Ward criterion  ClustGeo proposed in \cite{chavent2018clustgeo}. We empirically find that combining these two kinds of distances by computing the average of their respective standardized versions consistently achieves the best performance in different datasets, as shown in \Cref{tab:ablation-cluster-construction}.

\noindent \textbf{Number of clusters.} 
We also experiment with the number of clusters to generate for each dataset. In order to keep the performance of ELFIS consistent across different datasets, we define the number of clusters to generate as a fraction of the total number of classes of the dataset. We explore multiple ratios (5\%, 10\%, and 15\%) and show in \Cref{tab:ablation-num-clusters} that 10\% corresponds to a suitable tradeoff between classification performance and model complexity.
In this regard, we also compare the performance obtained using the proposed clusters and logical handcrafted partitions, such as the manufacturers of the planes in FGVC-Aircraft. The results in \Cref{tab:ablation-aircraft} indicate that class subsets are important in training the experts, and that model-centered grouping  considerably outperforms existing taxonomies.



\noindent \textbf{Knowledge distillation.} We perform a series of experiments in order to determine the influence of knowledge distillation/mutual learning on the performance of ELFIS, whose results are shown in \Cref{tab:ablation-mutual}. We can see that the best overall configuration in the two considered datasets is mutual learning, i.e., using the two terms of equation \eqref{eq:mutual-learning}. In the case of CUB-200-2011, we see a clear improvement of $+0.6$ in accuracy. These results highlight the benefits of using two-way distillation between the original classification head and the aggregated classifier, showing a much higher performance than the one-way distillation counterpart.

\noindent \textbf{Expert size.} We also explore the possibility of increasing or decreasing the size of the replicated expert heads, in an attempt of finding a suitable tradeoff between classification performance and model complexity. Particularly, using EfficientNet-B0 as a baseline model, we experiment with replicating different amounts of the last mobile inverted bottleneck MBConv \cite{sandler2018mobilenetv2, tan2019mnasnet} blocks that make up the EfficientNet architecture. More concretely, we test the influence of replicating either the last two MBConv blocks, only the last one, or none of them (just replicating the trailing convolutional layer for each head), and report the corresponding results in \Cref{tab:ablation-num-blocks}. We observe that, even though the best results are obtained by replicating the last MBConv block, not replicating any block still leads to results that outperform the original baseline. This  hints that most of the improvements provided by ELFIS are associated with the backbone learning more discriminative features, rather than just attributing the improvement to increased model complexity.

\subsection{Qualitative results}

In order to better understand and visualize the behavior of ELFIS, we perform a series of qualitative analyses, such as visualizing with GradCam \cite{selvaraju2017grad} the regions of the input image which are the most relevant for the model's final predictions, as well as visualizing in two dimensions the model's internal image representations using UMAP \cite{mcinnes2018umap}.

\begin{figure*}
\begin{center}
\includegraphics[width=0.95\linewidth]{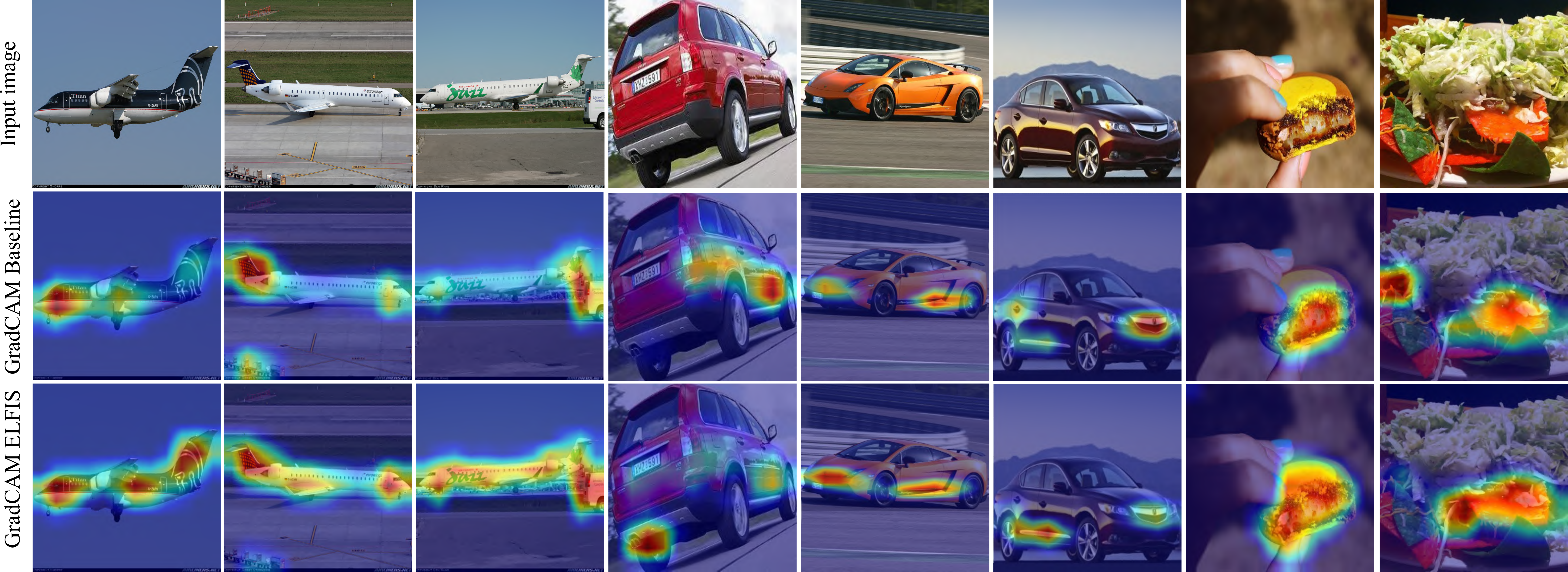}
\end{center}
   \caption{Examples of images from FGVC-Aircraft, Stanford Cars, and Food-101 in which the baseline fails and ELFIS predicts the right class. The corresponding GradCAM outputs are shown to highlight the differences between both models.}
\label{fig:gradcam}
\end{figure*}

\begin{figure}[t]
    \centering
    \includegraphics[width=\linewidth]{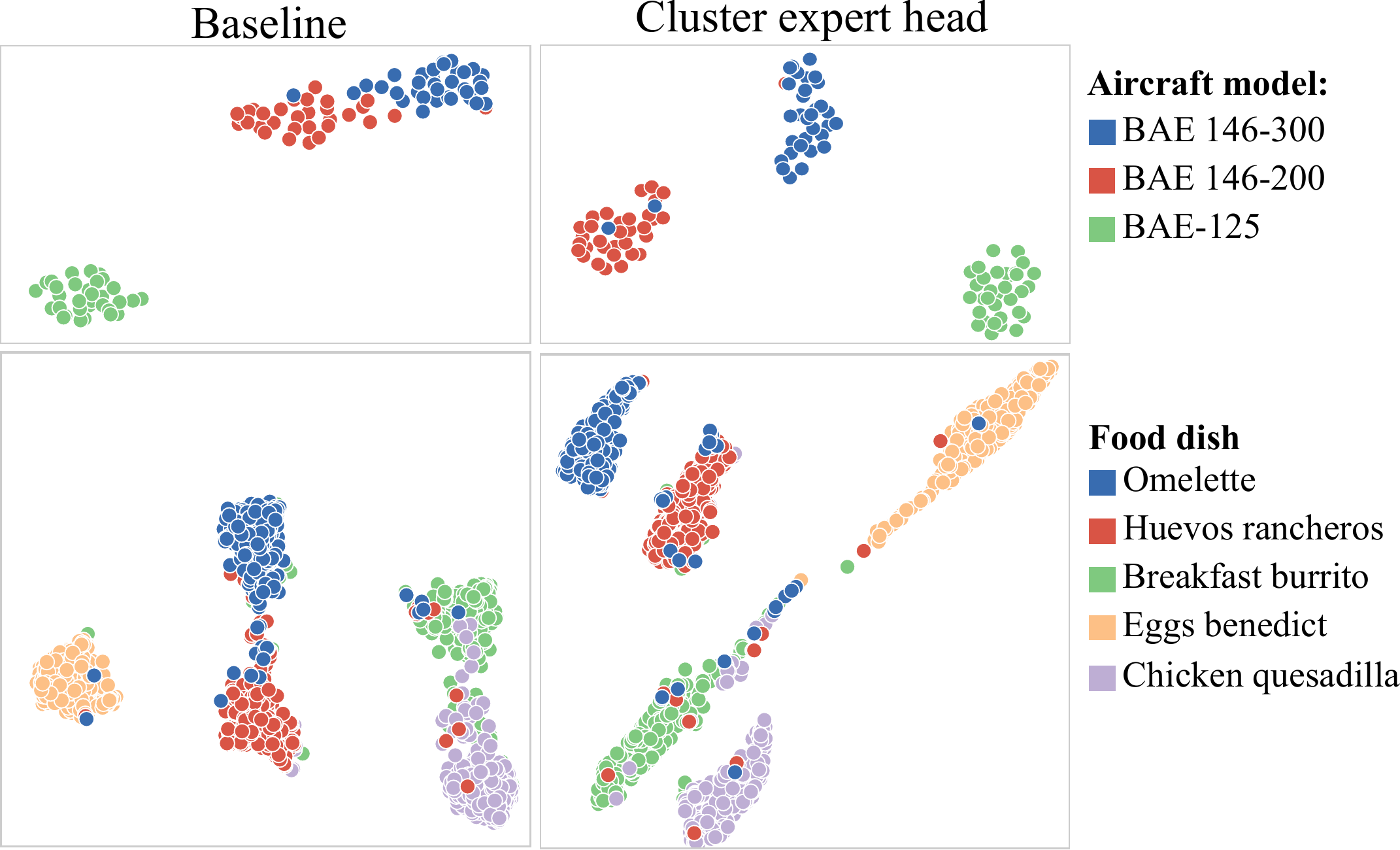}
    \caption{UMAP of the embeddings of the baseline and the expert heads for 2 clusters from Aircraft using EfficientNet-B0 (first row) and Food-101 with ResNet-50 (second row).}
    \label{fig:umaps}
\end{figure}

In \Cref{fig:gradcam}, we display some GradCAM \cite{selvaraju2017grad} examples for different datasets, showing in each column images where, unlike the baseline, ELFIS is capable of correctly predicting the class label. We find that ELFIS is capable of better identifying patterns that allow for the fine-grained discrimination of visually very similar object classes. As shown in \Cref{fig:gradcam}, ELFIS is capable of correctly distinguishing similar airplane models based on their length, similar car models based on subtle details such as having a double exhaust pipe or some other fine details, as well as closely related food dishes by correctly paying attention to the most discriminative ingredient. Our GradCAM visualizations show that, unlike the baseline model, ELFIS is indeed focusing on all these discriminative regions of the image, resulting in  a correct prediction.

To assess whether ELFIS is helping the network learn more discriminative representations for the fine-grained images, we visualize in 2D the intermediate features generated by both the baseline model and ELFIS. More concretely, in \Cref{fig:umaps} we visualize, for two different datasets, the zoomed-in UMAP representations of the images of a certain cluster generated by the original baseline and the corresponding subset-specific head from ELFIS. We find that ELFIS is capable of better separating the different fine-grained classes within the subset in the feature space, which translates into a direct improvement of the classification performance. This finding goes in line with those reported in the analysis of the GradCAM outputs in \Cref{fig:gradcam}. More UMAP examples are included in Appendix B, and more examples in which ELFIS outputs a better prediction can be found in Appendix C.




\subsection{Limitations}
Despite the excellent results, there are some limitations that might affect the benefits of using ELFIS. (1) The \textbf{optimality of the generated clusters} directly depends on the quality of the lexical embedding. Although we have tested our framework against a variety of domains, it is possible that in some very specific datasets general language models might not represent the real similarity between class labels, which might undermine the final performance of the model.
(2) Although the results of our ablations show strong improvements when using very reduced experts, it must be considered that their addition involves some \textbf{overhead} with respect to the baseline. This might affect its applicability to some time-sensitive tasks.
(3) It must be considered that our clustering assumes a sparse confusion matrix, having only tested our proposal on already highly performant baselines. It is yet to be explored whether ELFIS will achieve comparable improvements when built on top of \textbf{weaker baselines}.

\section{Conclusions}

In this paper, we introduce ELFIS, an 
expert learning framework for FGVR based on subset learning. 
We propose a method that defines subsets of categories of a dataset based on the 
visual similarity
of any given baseline network, as well as lexical information extracted from the dataset itself. We also introduce an effective method for exploiting this information. As a result, ELFIS acts as a universal add-on that enhances the classification ability of any deep neural network, emphasizing on the performance of the most challenging classes for the given baseline model.
The obtained results show the great versatility of ELFIS, as well as its capability to obtain competitive results with current SoTA approaches for a variety of public FGVR benchmarks. Our thorough analysis verifies that the proposed method enables the learning of more useful features to discriminate highly similar classes.
Regarding future lines of work, 
it is a natural question whether the proposed grouping technique can be adapted to build  multi-level hierarchies and how this additional information can be exploited to further enhance the capabilities of ELFIS.

\bibliographystyle{abbrv}
\bibliography{refs}

\section*{Appendix}
\section*{A. Learning curves}\label{sec:curves}

\begin{figure}[!ht]
    \centering
    \includegraphics[width=\linewidth]{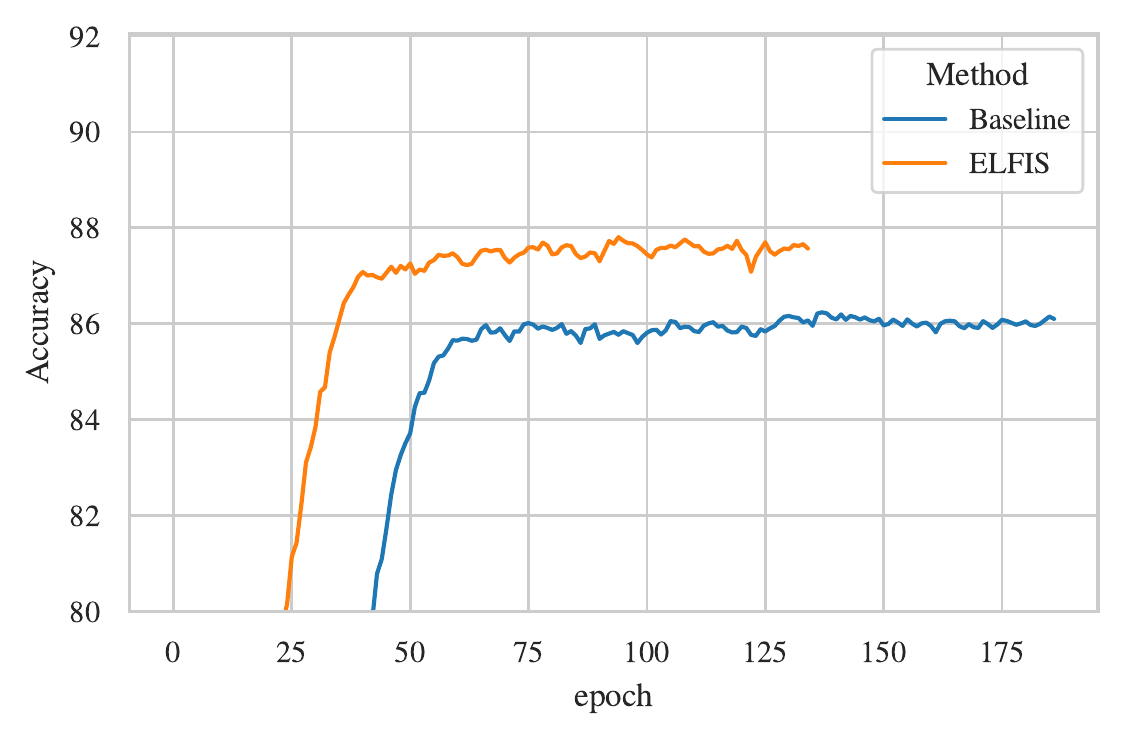}
    \caption{Accuracy of ResNet-50 for CUB-200-2011.}
    \label{fig:cub_resnet}
\end{figure}

\begin{figure}[!ht]
    \centering
    \includegraphics[width=\linewidth]{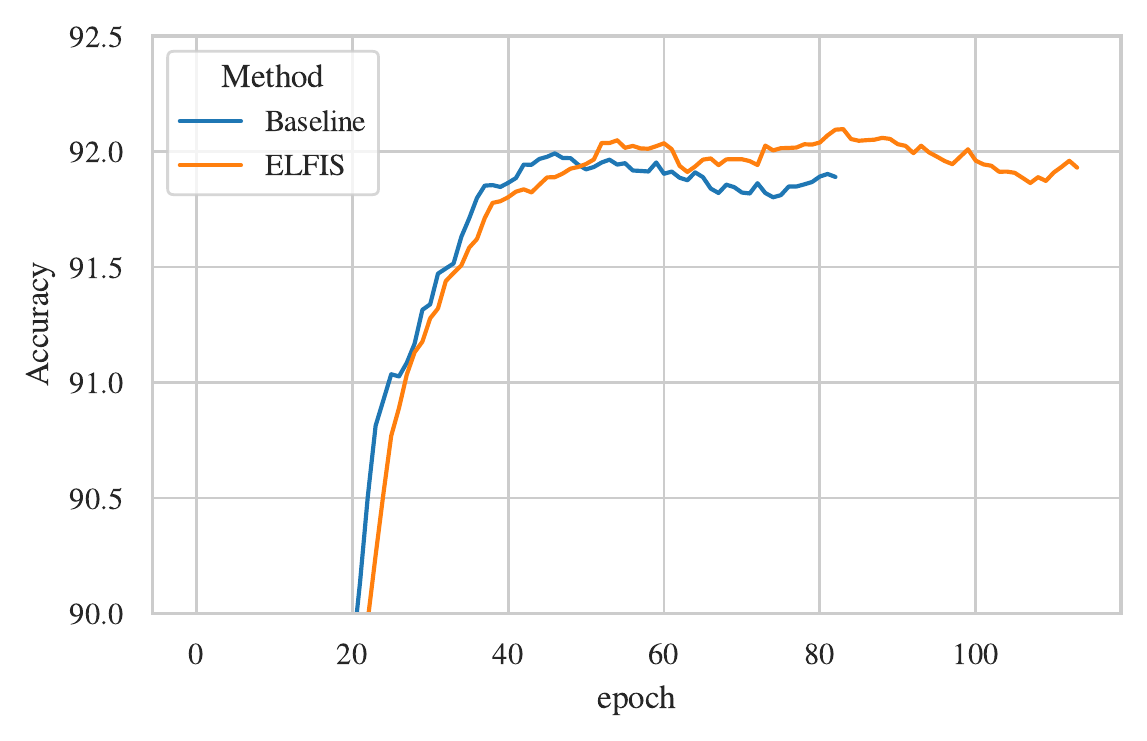}
    \caption{Accuracy of Swinv2-B for CUB-200-2011.}
    \label{fig:cub_swin}
\end{figure}

\begin{figure}[!ht]
    \centering
    \includegraphics[width=\linewidth]{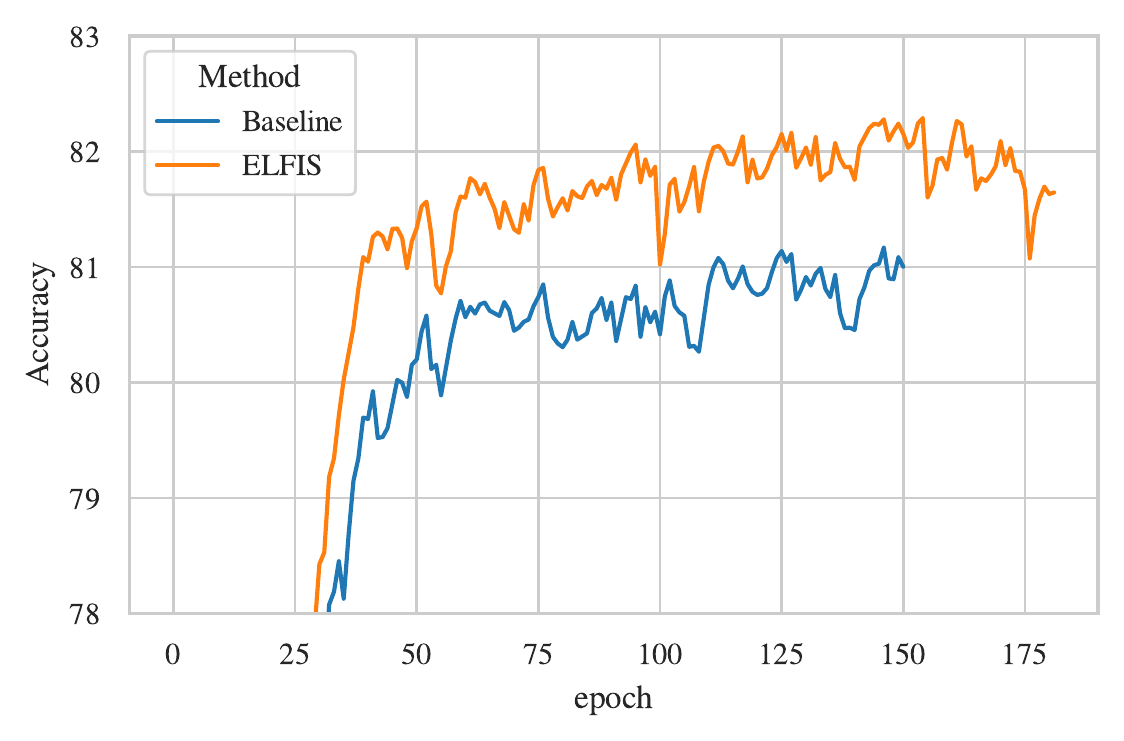}
    \caption{Accuracy of EfficientNet-B0 for NABirds.}
    \label{fig:nab_effnet}
\end{figure}

\begin{figure}[!ht]
    \centering
    \includegraphics[width=\linewidth]{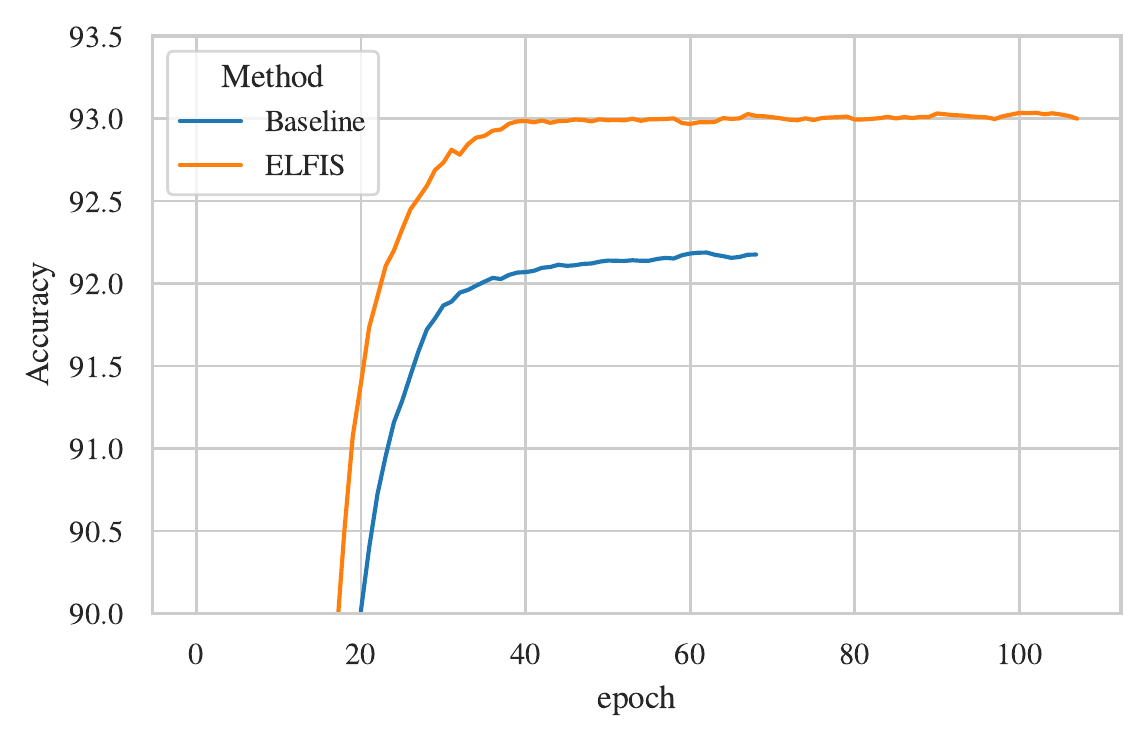}
    \caption{Accuracy of ViT-B/16 for Food-101.}
    \label{fig:f101_vit}
\end{figure}

In this section, we compare the evolution of accuracy over time for both the baseline and ELFIS. More concretely, in \Cref{fig:cub_resnet,fig:cub_swin,fig:nab_effnet,fig:f101_vit} we explore one example for each one of the studied backbones: EfficientNet-B0, ResNet-50, ViT-B/16 and Swinv2-B. Please note that, in all the cases, the curves have been smoothed to make them easier to visualize, and keep in mind that, as explained in the main text, early-stopping was used in all the experiments.

\Cref{fig:cub_resnet} contains the accuracy of ResNet-50 when training on CUB-200-2011. We can observe that ELFIS outperforms the baseline throughout the whole training process. Also, it can be seen that it reaches results comparable to the best baseline results much earlier in the training.

Also for CUB-200-2011, we display in \Cref{fig:cub_swin} the accuracy curves of Swinv2-B. Note, here we zoom in on the vertical axis to allow a clear distinction between the lines. As the baseline is much stronger than the case of ResNet-50, surpassing it is more difficult for ELFIS in this case, and it needs around 50 epochs to achieve higher accuracy. Nevertheless, at the end of the training, ELFIS is consistently better than the baseline.

Regarding EficientNet-B0, \Cref{fig:nab_effnet} shows how for NABirds ELFIS outperforms the baseline in all the epochs, as it was the case in ResNet-50. 

Finally, we show in \Cref{fig:f101_vit} a zoomed-in version of the accuracy curves of ViT-B/16 for Food-101. Again, the difference between the baseline and ELFIS is noticeable and consistent during the training process.

As we have seen, ELFIS is able to obtain clear and stable gains in accuracy over the baseline regardless of the backbone used and the capacity. This finding goes in line with the results discussed in the main text and reinforces the idea that ELFIS is able to help improve already top-performant solutions. 

\section*{B. Feature representations}

\begin{figure*}
\floatsetup{valign=t, heightadjust=all}
    \centering
\ffigbox{
    \begin{subfloatrow}
    \centering
    \ffigbox{\includegraphics[width=\linewidth]{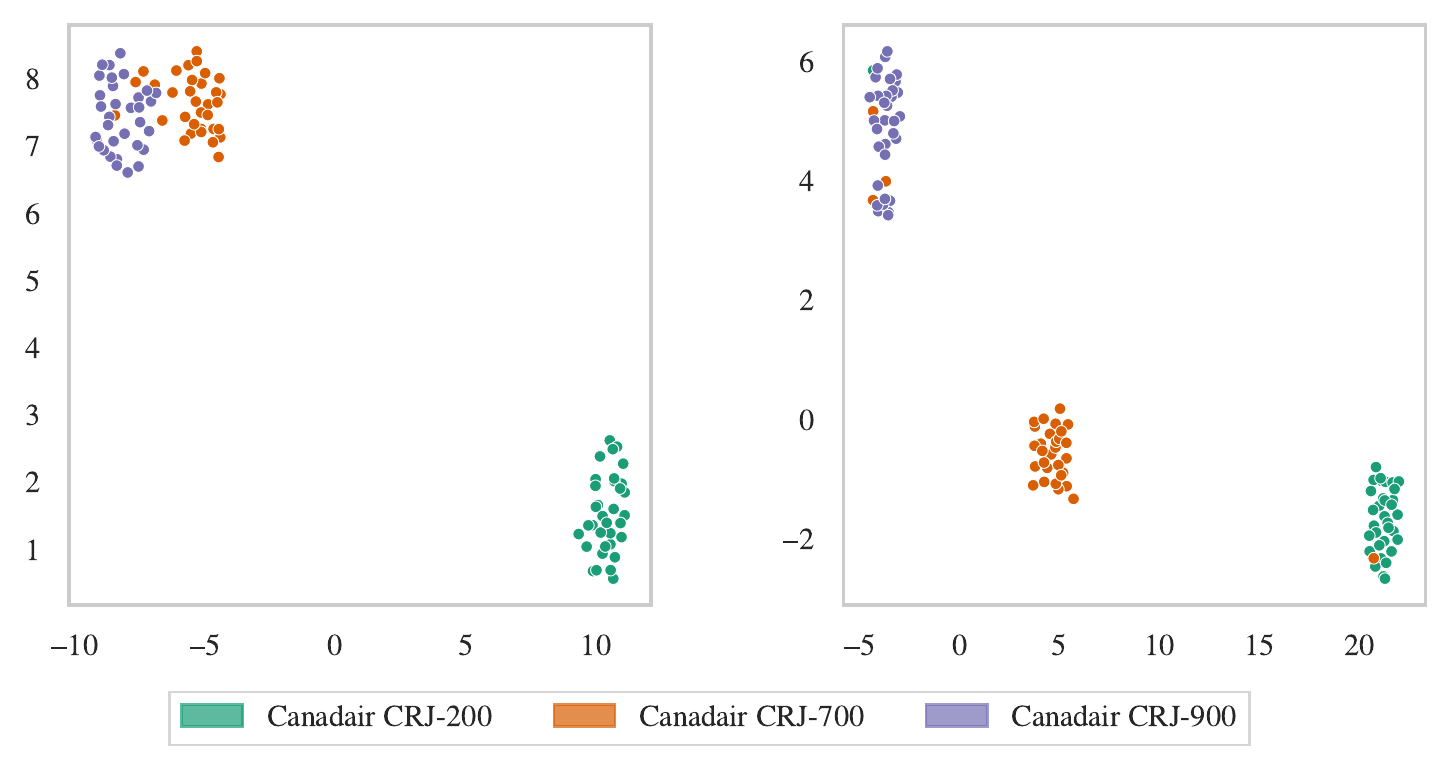}}{\caption{FGVC-Aircraft with EfficientNet-B0. Cluster 9. \label{fig:effb0_aircraft_9}}}
    \ffigbox{\includegraphics[width=\linewidth]{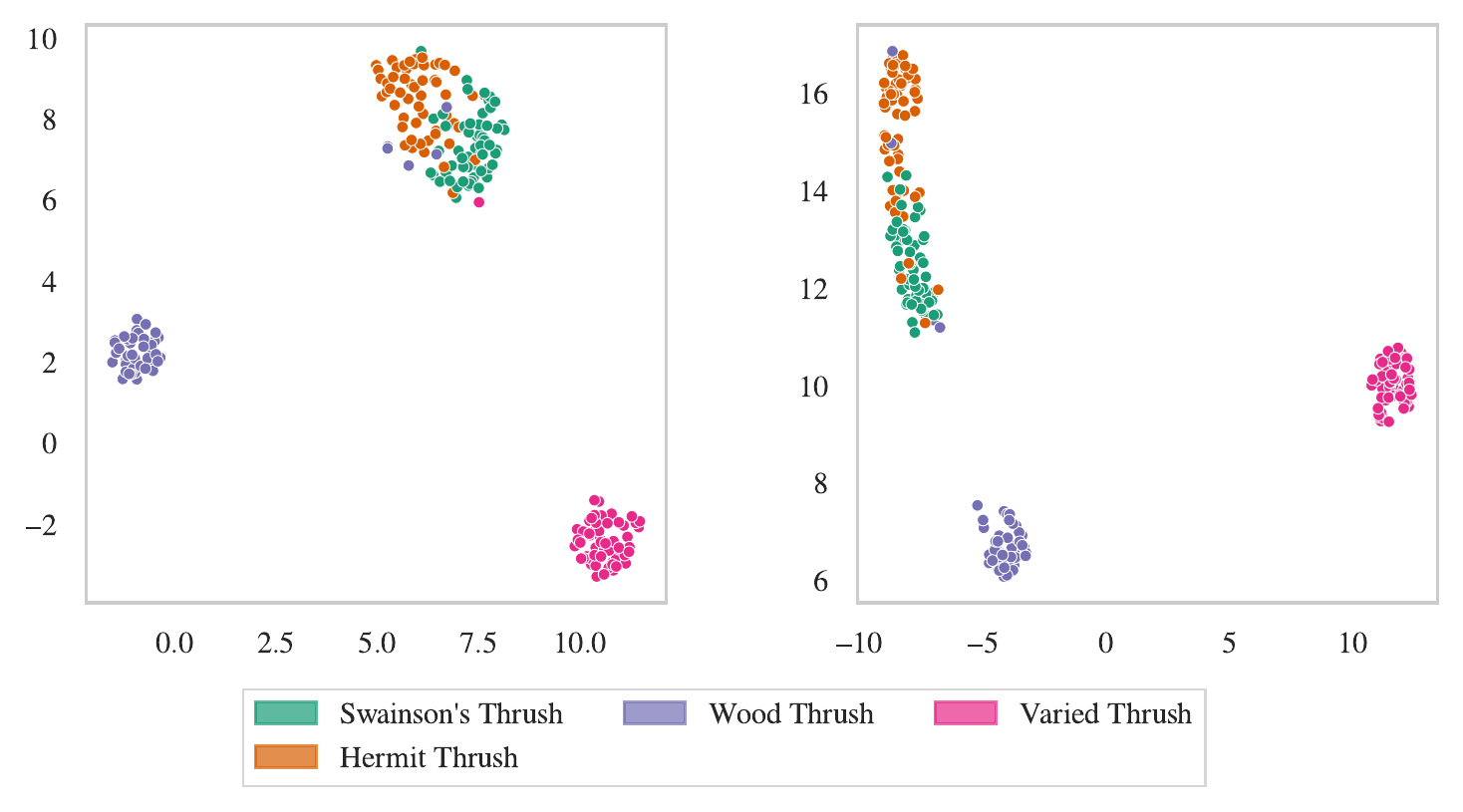}}{\caption{NABirds with EfficientNet-B0. Cluster 4. \label{fig:effb0_nabirds_4}}}
    \end{subfloatrow}

    \begin{subfloatrow}
    \centering
    \ffigbox{\includegraphics[width=\linewidth]{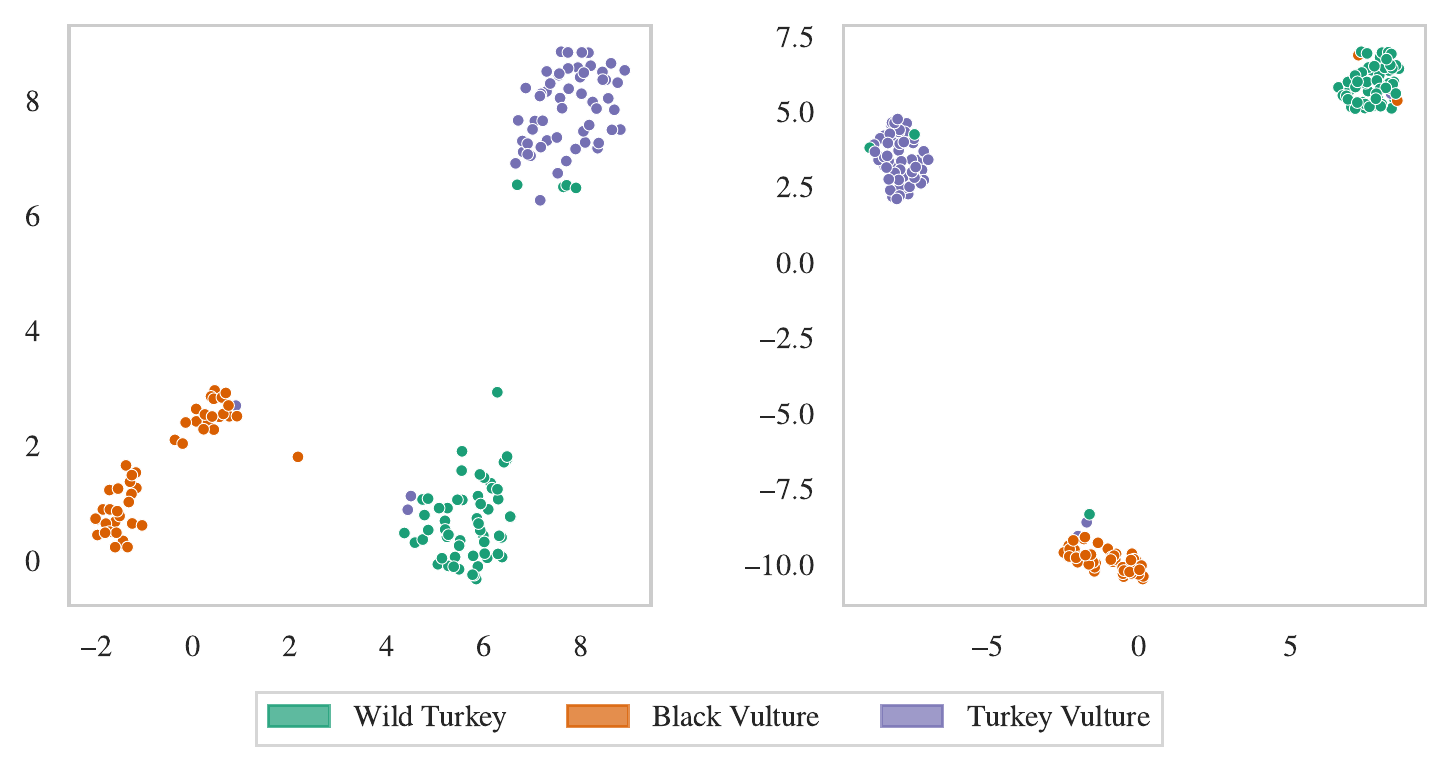}}{\caption{NABirds with EfficientNet-B0. Cluster 7. \label{fig:effb0_nabirds_7}}}
    \ffigbox{\includegraphics[width=\linewidth]{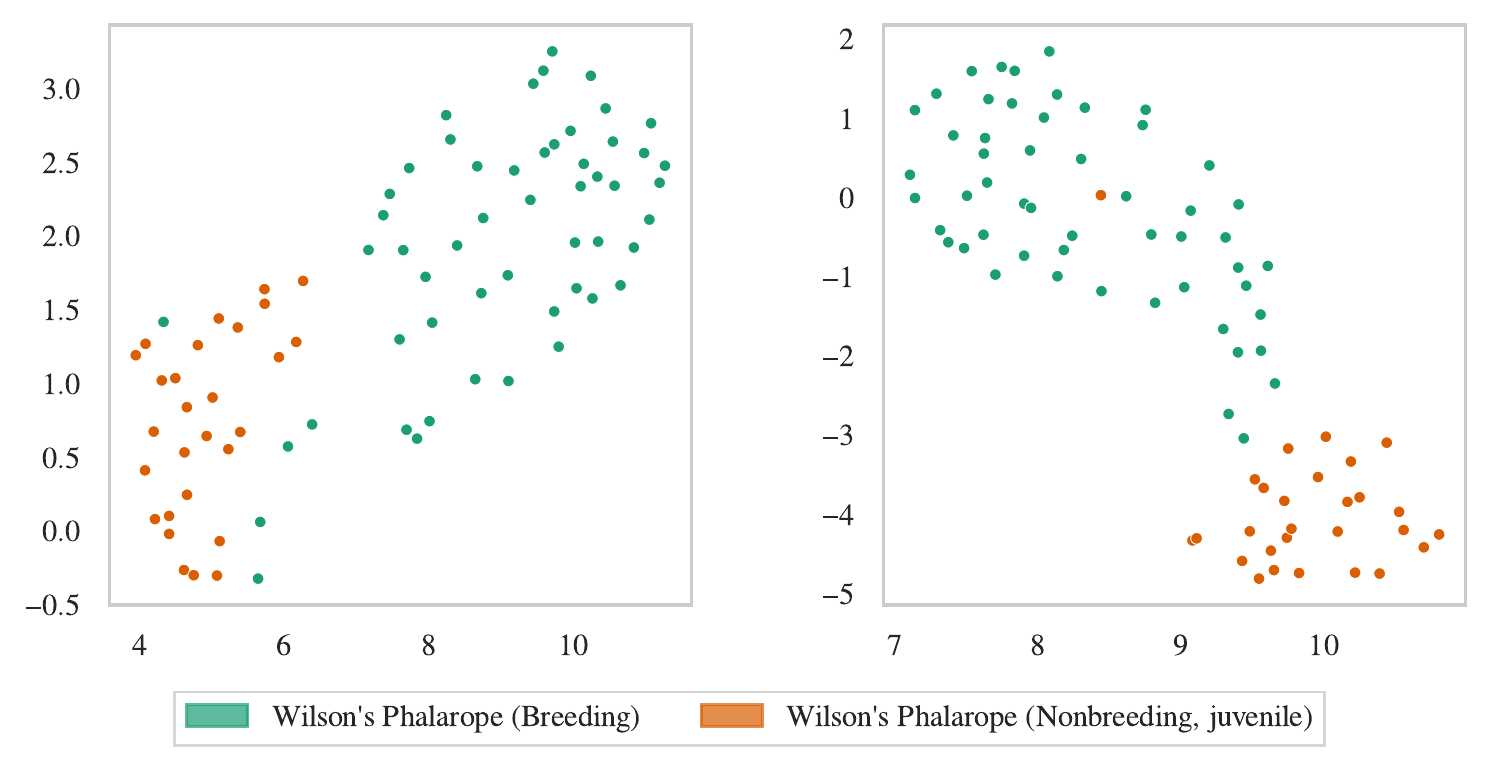}}{\caption{NABirds with EfficientNet-B0. Cluster 11. \label{fig:effb0_nabirds_11}}}
    \end{subfloatrow}

    \begin{subfloatrow}
    \centering
    \ffigbox{\includegraphics[width=\linewidth]{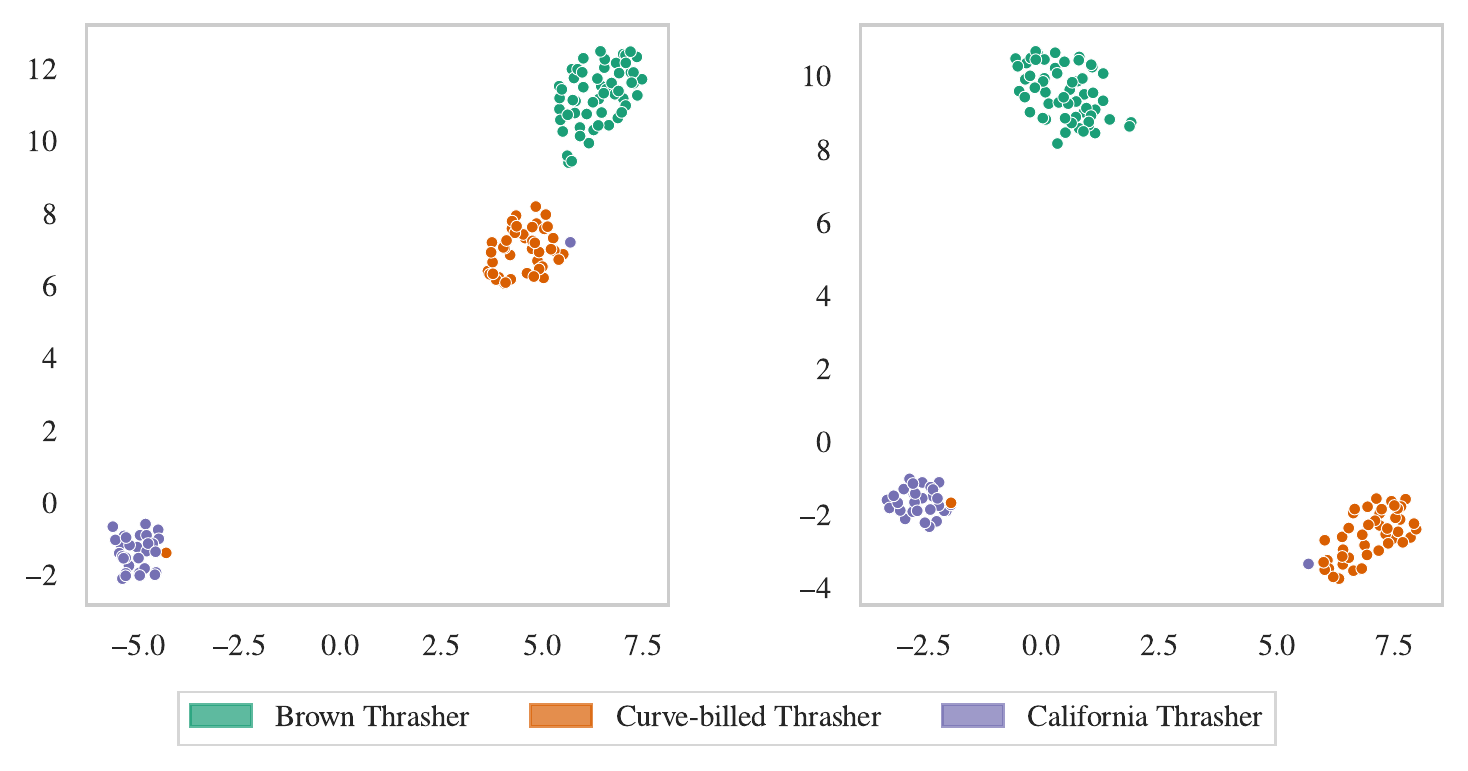}}{\caption{NABirds with EfficientNet-B0. Cluster 12. \label{fig:effb0_nabirds_12}}}
    \ffigbox{\includegraphics[width=\linewidth]{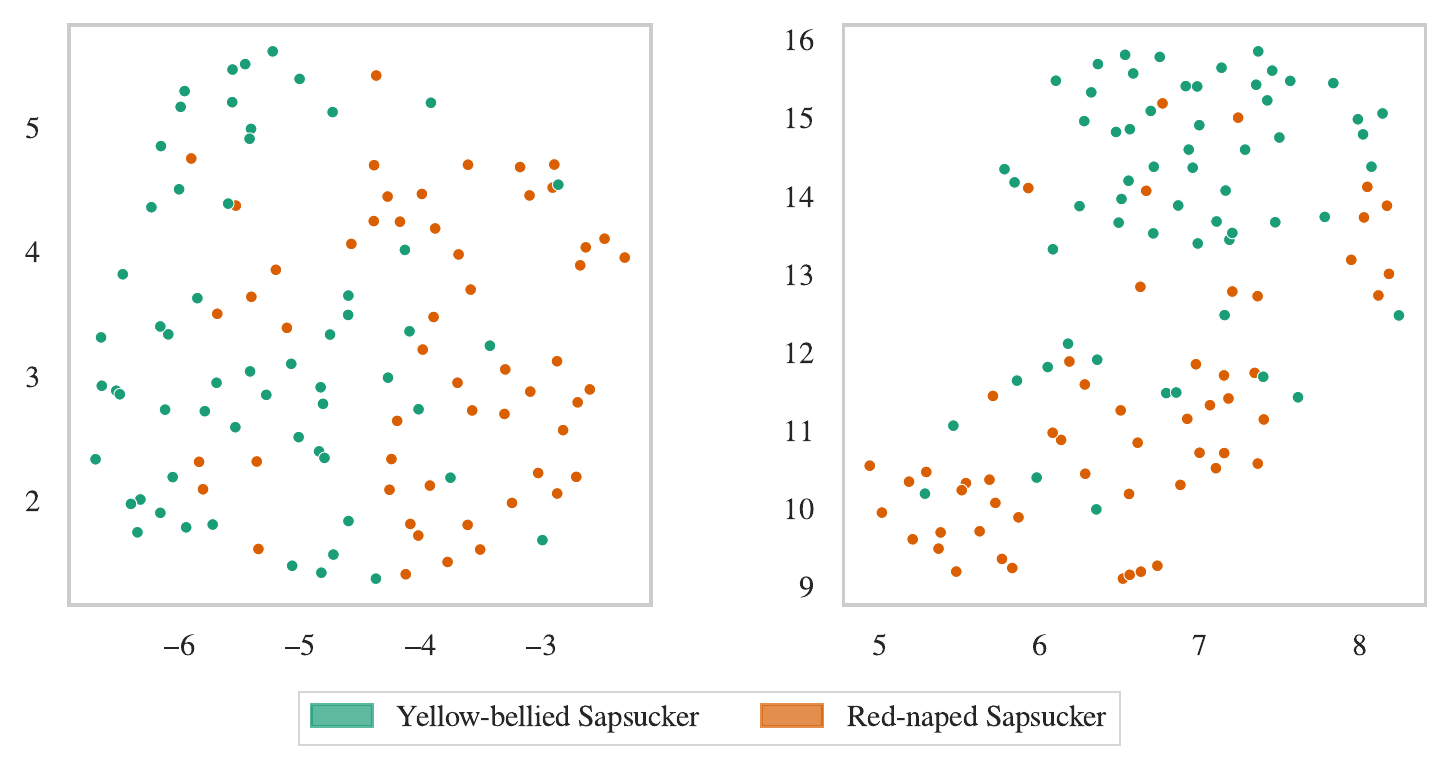}}{\caption{NABirds with EfficientNet-B0. Cluster 18. \label{fig:effb0_nabirds_18}}}
    \end{subfloatrow}

    \begin{subfloatrow}
    \centering
    \ffigbox{\includegraphics[width=\linewidth]{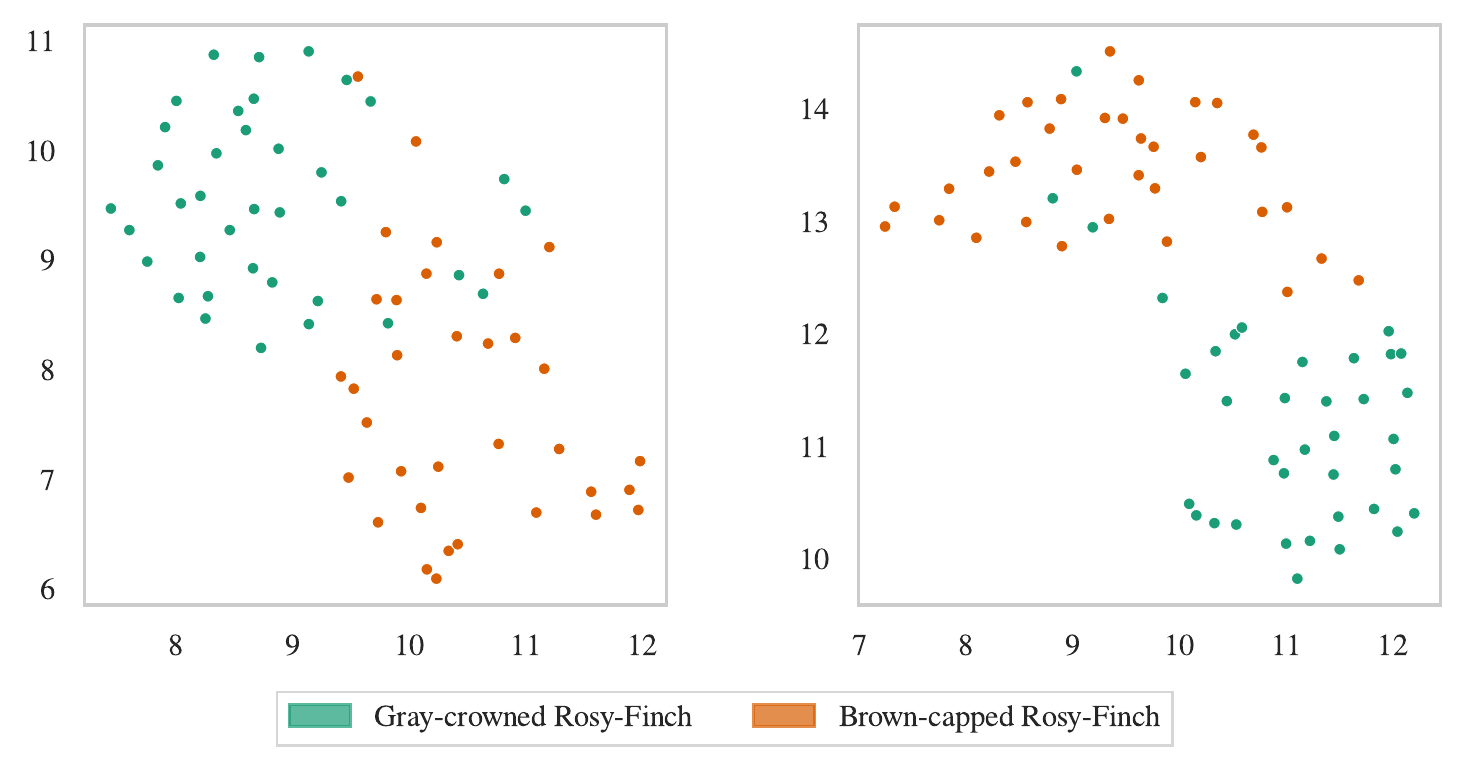}}{\caption{NABirds with EfficientNet-B0. Cluster 24. \label{fig:effb0_nabirds_24}}}
    \ffigbox{\includegraphics[width=\linewidth]{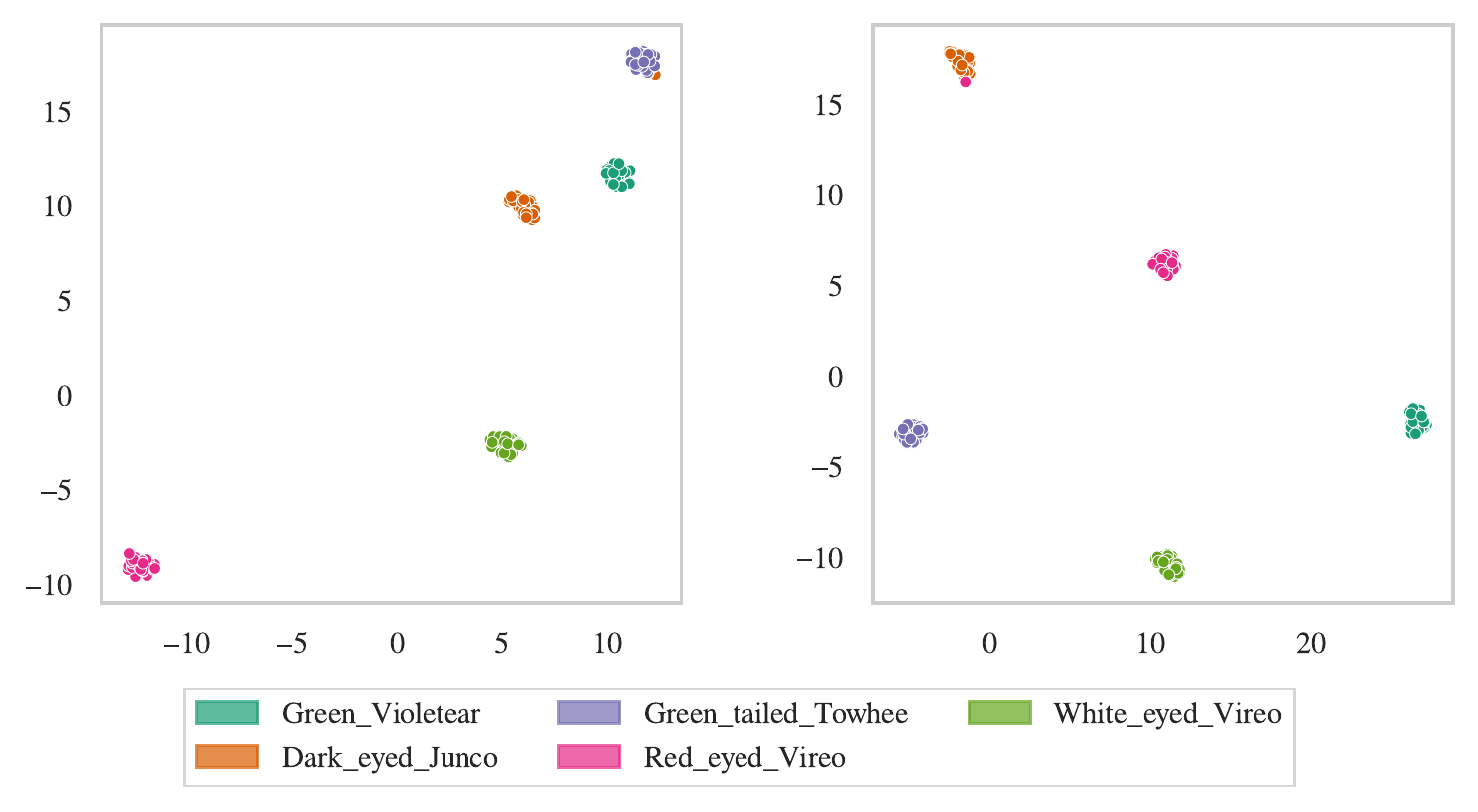}}{\caption{CUB-200-2011 with Swinv2-B. Cluster 10. \label{fig:swin_cub_10}}}
    \end{subfloatrow}
}
{\caption{Examples of UMAP embeddings (I). In each subfigure, the UMAP on the left corresponds to the baseline and the UMAP on the right to the expert head of ELFIS.}\label{fig:umap1}}
\end{figure*}

\begin{figure*}
\floatsetup{valign=t, heightadjust=all}
    \centering
\ffigbox{
    \begin{subfloatrow}
    \centering
    \ffigbox{\includegraphics[width=\linewidth]{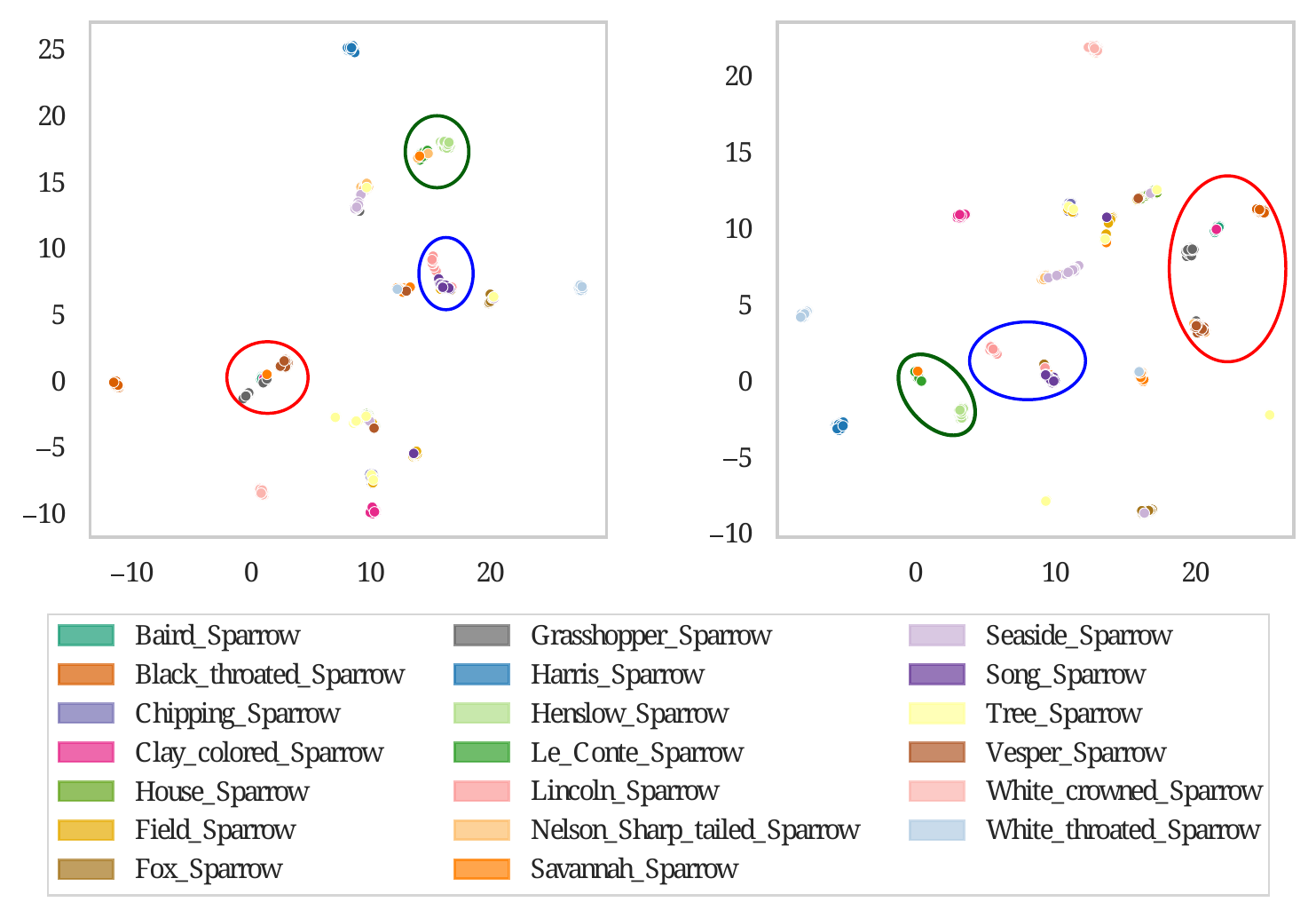}}{\caption{CUB-200-2011 with Swinv2-B. Cluster 6. \label{fig:swin_cub_6}}}
    \ffigbox{\includegraphics[width=\linewidth]{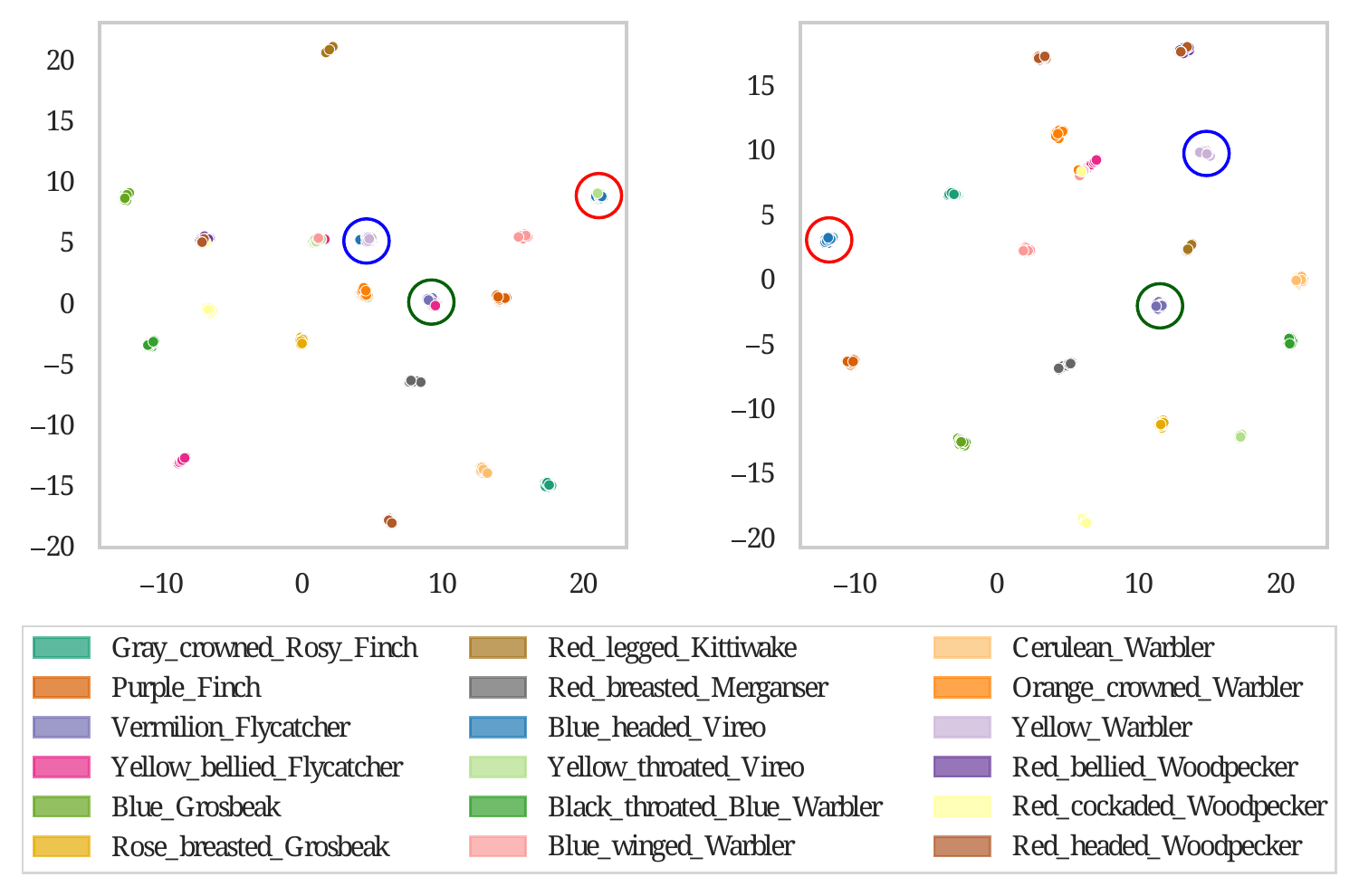}}{\caption{CUB-200-2011 with Swinv2-B. Cluster 2. \label{fig:swin_cub_2}}}
    \end{subfloatrow}

    \begin{subfloatrow}
    \centering
    \ffigbox{\includegraphics[width=\linewidth]{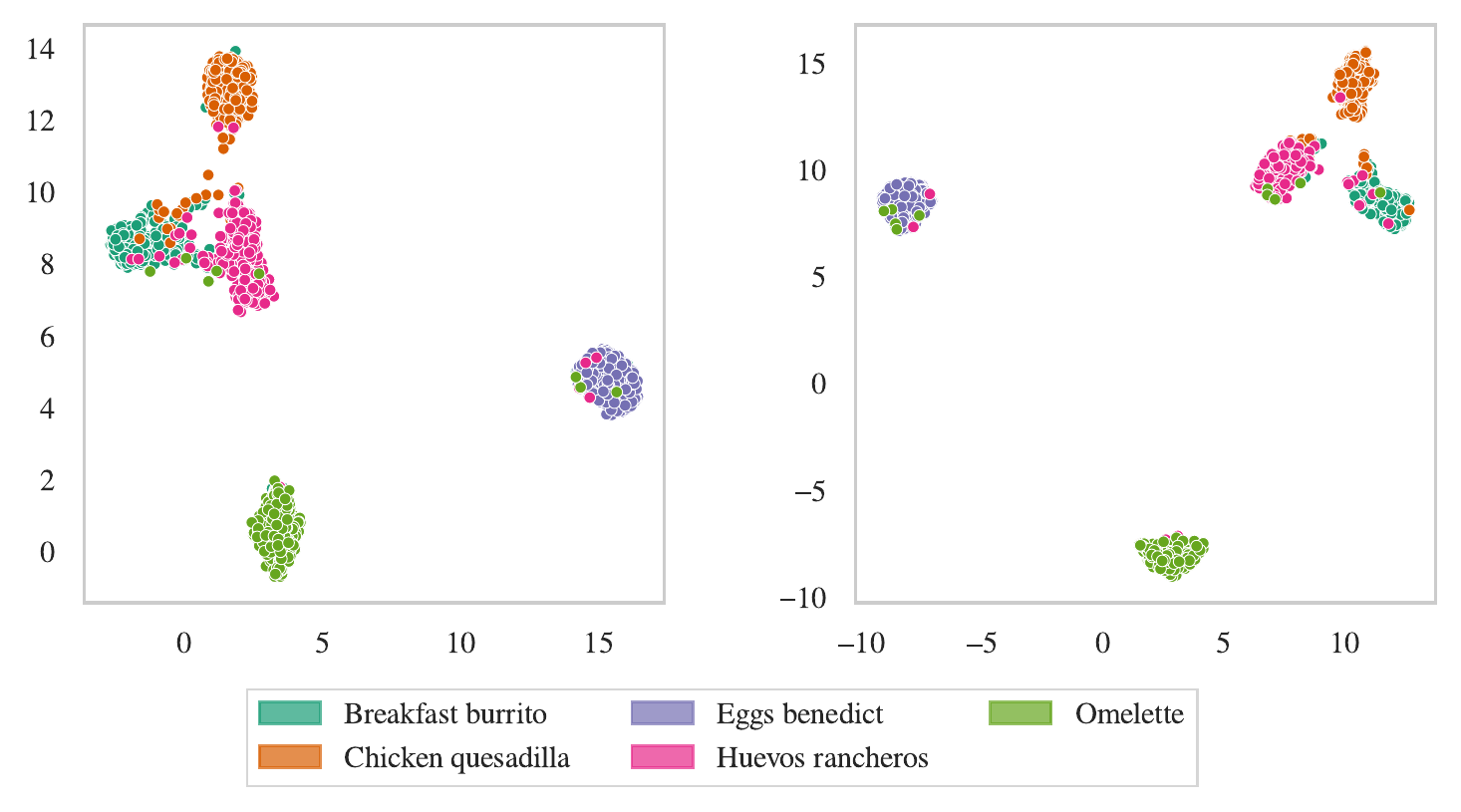}}{\caption{Food-101 with ViT-B/16. Cluster 4. \label{fig:vit_food_4}}}
    \ffigbox{\includegraphics[width=\linewidth]{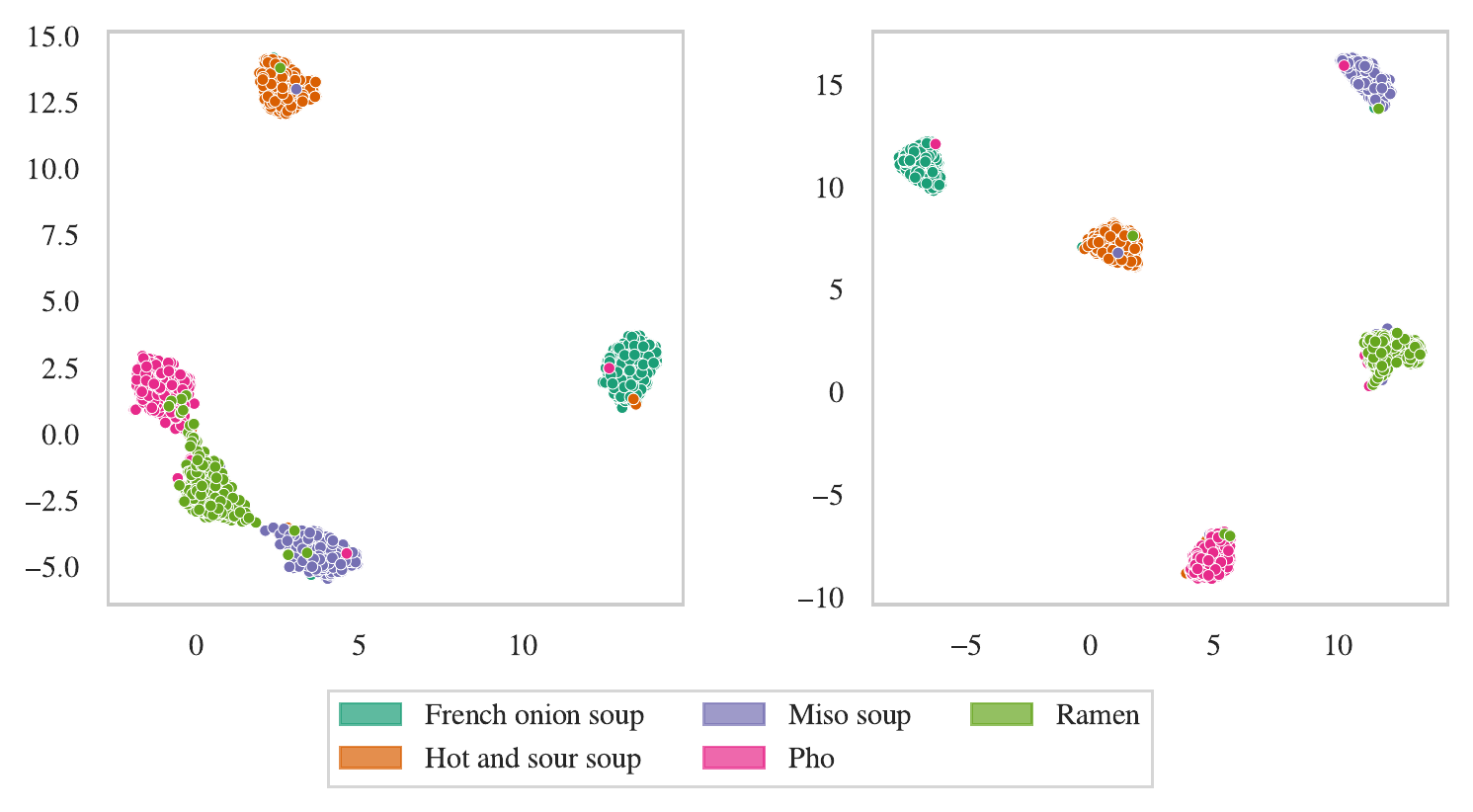}}{\caption{Food-101 with ViT-B/16. Cluster 1. \label{fig:vit_food_1}}}
    \end{subfloatrow}

    \begin{subfloatrow}
    \centering
    \ffigbox{\includegraphics[width=\linewidth]{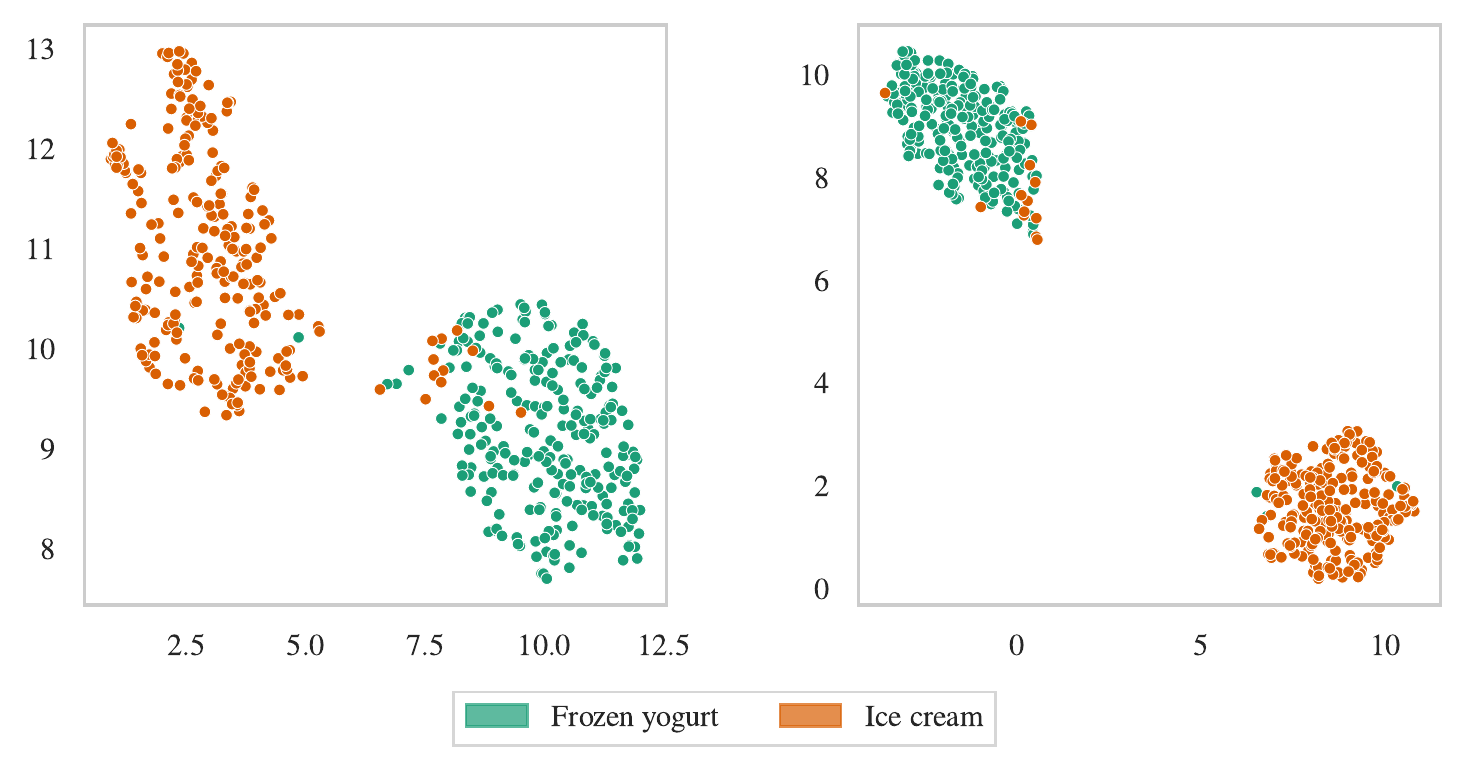}}{\caption{Food-101 with ViT-B/16. Cluster 6. \label{fig:vit_food_6}}}
    \ffigbox{\includegraphics[width=\linewidth]{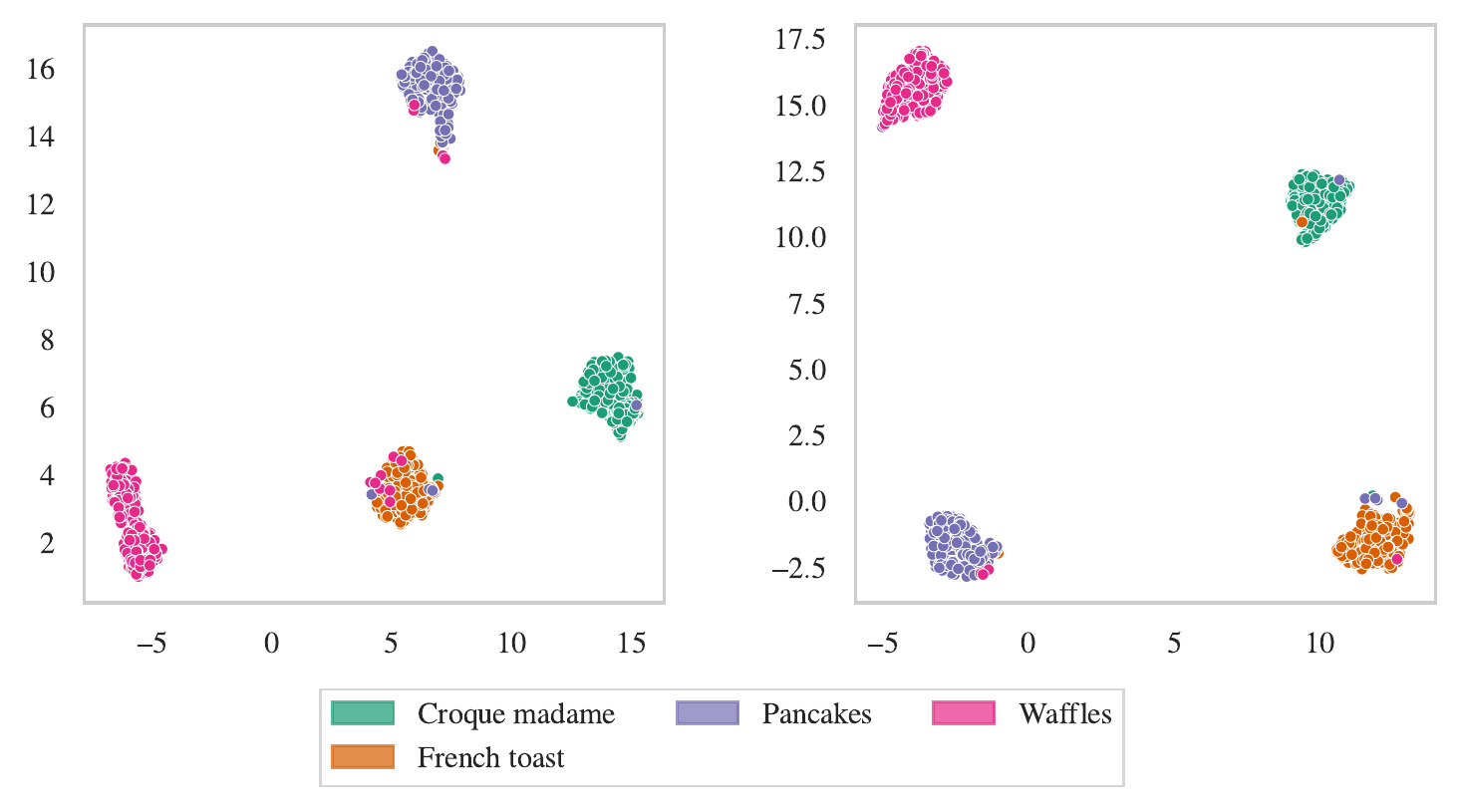}}{\caption{Food-101 with ViT-B/16. Cluster 2. \label{fig:vit_food_2}}}
    \end{subfloatrow}
}
{\caption{Examples of UMAP embeddings (II). In each subfigure, the UMAP on the left corresponds to the baseline and the UMAP on the right to the expert head of ELFIS.}\label{fig:umap2}}
\end{figure*}

In addition to the two UMAP examples that were included in the main text, we display in \Cref{fig:umap1,fig:umap2} more examples in which including ELFIS helps the model learning better feature representations to distinguish classes within the same cluster. 

In \Cref{fig:effb0_aircraft_9} we show another Aircraft example very similar to the one we introduced in the main text. As we can clearly see, this cluster is formed by three planes of the same brand, and while the baseline struggles to tell them apart, ELFIS can easily find distinguishable representations in the feature space.

\Cref{fig:effb0_nabirds_4,fig:effb0_nabirds_7,fig:effb0_nabirds_11,fig:effb0_nabirds_12,fig:effb0_nabirds_18,fig:effb0_nabirds_24} contain examples of EfficientNet-B0 with the NABirds dataset. In some of the cases, like \Cref{fig:effb0_nabirds_7,fig:effb0_nabirds_12} the representations of the different classes are clearly better separated in ELFIS. In others, such as \Cref{fig:effb0_nabirds_11,fig:effb0_nabirds_18,fig:effb0_nabirds_24}, there are two classes whose representations are highly entangled before using ELFIS, however, the degree of overlap is noticeably decreased by our proposed method. There is a similar situation in \Cref{fig:effb0_nabirds_4}, in which 2 of the 4 classes are slightly separated by ELFIS.

The next three images, \Cref{fig:swin_cub_10,fig:swin_cub_2,fig:swin_cub_6}, contain UMAP representations of the embeddings of Swinv2-B for CUB-200-2011. In this case, we show larger clusters, which evidences that ELFIS can work with not-so-specialized experts. In \Cref{fig:swin_cub_10}, we observe that the representations of the different classes are better spread in the feature space with ELFIS. In \Cref{fig:swin_cub_6,fig:swin_cub_2}, we visualize two clusters with a large number of classes. For each of them, we have highlighted different areas in which ELFIS is able to separate better similar classes or reduce the number of misclassifications. 

Finally, we display 4 examples of clusters using ViT-B/16 for the Food-101 dataset. \Cref{fig:vit_food_1,fig:vit_food_2,fig:vit_food_4,fig:vit_food_6} show clear improvements in the representations learned by the model when including ELFIS. The expert helps the network to spread the different classes in different areas of the feature space, making it easier to classify those classes. It can be noted that, in the case of \Cref{fig:vit_food_4}, the cluster and the results are very similar to those shown in the main text for Food-101 and ResNet-50.

The UMAPs of both the main document and the supplementary material
strengthens our claim on the versatility of ELFIS.

\section*{C. Particular examples} \label{sec:sup-particular-examples}

\begin{figure*}
    \includegraphics[width=\textwidth]{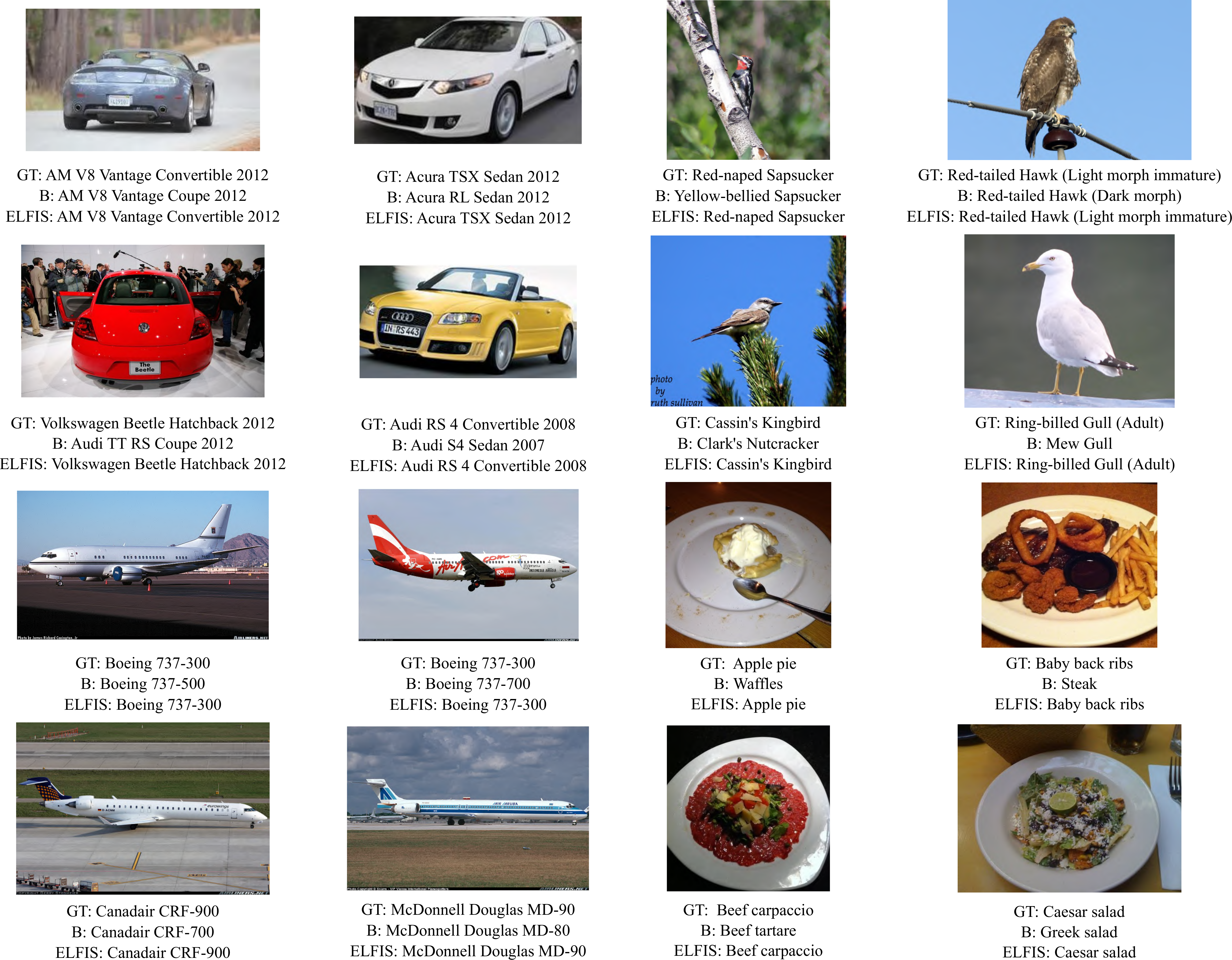}
    \caption{Examples of images in which the baseline predicts a wrong class and ELFIS makes a proper prediction. Images belong to Stanford Cars, NABirds, FGVC-Aircraft, and Food-101. All these examples were extracted using EfficientNet-B0 except for Food-101 (ResNet-50). Note that for each image we include: the ground truth class of the image (GT), the class wrongly predicted by the baseline (B), as well as the class predicted by ELFIS. }\label{fig:examples}
\end{figure*}

In \Cref{fig:examples}, we display some images of different datasets for which the baseline fails to find the proper class, while ELFIS succeeds. We attribute this improvement to the better ability of our method to separate the classes in the features space, as we show in the previous section.


\end{document}